# Multidimensional Bayesian Active Machine Learning of Working Memory Task Performance

Dom CP Marticorena[1], Chris Wissmann[1], Zeyu Lu[2], Dennis L Barbour[1,2]

[1]Department of Biomedical Engineering, Washington University, 1 Brookings Drive, St. Louis, MO 63130
[2]Department of Computer Science and Engineering, Washington University, 1 Brookings Drive, St. Louis, MO 63130

## Abstract

While adaptive experimental design has outgrown one-dimensional, staircase-based adaptations, most cognitive experiments still control a single factor and summarize performance with a scalar. In contrast, we validate here a Bayesian, two-axis, active-classification approach in an immersive virtual testing environment, using a 5×5 spatial working-memory reconstruction task as an application domain. Two variables are controlled: spatial load $L$ (number of occupied tiles) and feature-binding load $K$ (number of distinct colors) of items. Stimulus acquisition is guided by posterior uncertainty of a nonparametric Gaussian Process (GP) probabilistic classifier, which outputs a surface over ($L$, $K$) rather than a single threshold or maximum span value. In a young adult population, we compare GP-driven Adaptive Mode (AM) with a traditional adaptive staircase Classic Mode (CM), which varies $L$ only at $K = 3$. Parity between the methods is achieved for this cohort, with an intraclass correlation coefficient of 0.755 at $K = 3$. As a secondary exploratory result, AM reveals individual differences in interactions between spatial load and feature binding. AM estimates converge more quickly than other sampling strategies, demonstrating that only about 30 samples are required for accurate fitting of the full model.

# Introduction

The simplest procedure for evaluating perception or cognition with accuracy-based queries is delivering the same stimuli or test items to all members of an experimental cohort in randomized order. Assuming no floor or ceiling effects are observed (i.e., task difficulty falls within the linear range of the psychometric curve for all participants), such a procedure can be readily interpretable with simple methods for both experimental and individual differences research. This assumption is rarely met, however, with realistic sample sizes and typical cohort heterogeneity. A straightforward compensatory strategy might be to increase the set size and range of test item difficulties to ensure that some subset is informative for each participant. This approach retains the use of simple analytical procedures but is grossly inefficient, leading to fatiguing test sessions that also tend to be underpowered, particularly for individual differences research (Haaf & Rouder, 2018; Rouder & Haaf, 2021). In order to retain practical testing lengths, cognitive function estimation therefore often varies only one dimension of task difficulty while holding other potentially relevant dimensions fixed: change a single factor (e.g., salience or sequence length or set size) and summarize performance with a single number such as maximum memory span or *d*-prime (Hautus et al., 2021; Kessels et al., 2000; Luck & Vogel, 1997).

Dynamically adapting task difficulty for each participant is a long-standing strategy to accelerate testing by discarding task items that are less informative for a particular participant in favor of items that are more informative (Leek, 2001; Levitt, 1971; Rose et al., 1970; Watson & Pelli, 1983). Delivering easier items after a fail and harder items after a pass in a predetermined staircase procedure is the classic way of achieving this goal. Such methods are effective but can be improved by quantifying the information provided from multiple previous items for selecting the most informative next item. This general approach falls under the category of adaptive Bayesian estimation and has been used to improve sampling procedures for estimating individual psychometric curves (King-Smith et al., 1994; Kontsevich & Tyler, 1999; Kujala & Lukka, 2006; Lesmes et al., 2006; Remus & Collins, 2008; Shen & Richards, 2013).

Multidimensional objective functions are more challenging to support with adaptive Bayesian updates, but the efficiency gains and the inference that can be obtained per unit time are generally higher in these cases, often substantially so (Cavagnaro et al., 2010; Kujala & Lukka, 2006; Marticorena, Wong, Browning, Wilbur, Davey, et al., 2024; Myung et al., 2013). This approach has been validated across cognitive domains by multiple research groups. (Chang et al., 2021) showed that Gaussian process (GP)-based classification can adaptively optimize a decision-making task with two design variables and recover nonlinear isocontours, and (S. H. Lee et al., 2022) used GP regression and experimental design optimization to extend a numerical-representation function from a conventional 1D design space to a 2D space. Typically, these complex functions require a defined, closed parametric form to perform the necessary computations for Bayesian updating and to achieve adequate efficiency gains, thus limiting the approach to scenarios of strong theoretical or empirical foundations for the regression model.

We have shown for a variety of perceptual and cognitive models, however, that highly flexible nonparametric or hierarchical machine learning approaches can be coupled with Bayesian active learning to efficiently construct item-level modeling of individual behavioral responses (Barbour et al., 2018, 2019; Kasumba et al., 2025; Marticorena, Wong, Browning, Wilbur, Davey, et al., 2024; Marticorena, Wong, Browning, Wilbur, Jayakumar, et al., 2024; Song et al., 2015; Song, Sukesan, et al., 2017). This approach holds great potential for extending individualized behavioral models in complex, multidimensional cognitive domains lacking strong prior beliefs for particular model forms. A critical test of nonparametric multidimensional psychometric assessment is therefore whether it can preserve the inference obtained from a conventional one-dimensional test while efficiently characterizing performance across additional task dimensions. Working memory (WM) provides a well-suited cognitive domain for such a test.

WM is considered fundamental to reasoning, language, and goal-directed action. Individual differences in WM forecast, for example, academic outcomes and vulnerability to neuropsychiatric and age-related decline (Alloway, Banner, et al., 2010; Alloway, Gathercole, et al., 2010; Engle, 2002; Just & Carpenter, 1992; Lee & Park, 2005; Park & Reuter-Lorenz, 2009; Peich et al., 2013). Classic span tasks, however, compress WM to a single “set-size” axis, conflating distinct mechanisms, such as maintaining bound feature-location representations and resisting interference, with capacity itself. For example, misbinding or “swap” errors are partly dissociable from complete forgetting (Baddeley, 2010, 2012; Bays et al., 2009; Ma et al., 2014; Parra et al., 2010), yet conventional accuracy-based assessments often collapse distinct failure modes, thus disregarding the information provided by error type.

To manipulate two interpretable contributors to WM difficulty separately, we designed a spatial working memory task with variable spatial load $L$ (number of occupied tiles) and feature-binding load $K$ (number of distinct colors). Participants’ attempts to reconstruct spatial patterns are used to estimate performance isocontours in the ($L, K$) domain rather than a single threshold. This multidimensional view preserves main effects and $L \times K$ interactions, yielding a more individualized diagnostic behavioral phenotype (Cavagnaro et al., 2010; Watson, 2017). Thus, the primary contribution of the present study is to validate a fast, nonparametric approach to multidimensional adaptive assessment that preserves conventional one-dimensional inference while efficiently characterizing additional task dimensions. The observed $L \times K$ individual differences illustrate the additional cognitive information made accessible by this approach.

We address the mismatch between one-factor staircases and multidimensional difficulty by validating a nonparametric Bayesian active-classification framework that learns each participant’s performance across the ($L, K$) domain. Using the predictive posterior from this sequential estimation process, we adaptively target the 50% performance isocontour, enabling direct comparison to a standard $L$-only staircase at $K = 3$. Conceptually, this procedure instantiates adaptive design optimization without committing to a closed-form psychometric function, thereby complementing QUEST+ when the link between task settings and performance is unknown or irregular (Myung et al., 2013; Watson, 2017).

The central contribution of this study is methodological rather than a new Gaussian-process formulation. We empirically validate a transferable multidimensional active-assessment procedure in a human cognitive task and quantify its sampling efficiency. The broader goal of this research is to demonstrate that optimal adaptive algorithms can allow considerably more complex behavioral models to be constructed for individual participants in comparable amounts of time as current reductionist test batteries. More contextualized multidimensional models can be deployed as a result. In this particular application, we test whether this procedure can recover the simpler inference of a basic spatial working memory task while also revealing more individual variation than is typically appreciated from such a task.

# Methods

## Overview

We implement a trial-based spatial working-memory reconstruction task, Build Master, in our custom immersive virtual behavioral testing platform pixelDOPA (Marticorena et al., 2025). Each trial has an Observation phase, in which participants reveal the colors of tiles on a 5×5 grid corresponding to a single test item, followed by a Build phase, in which they reconstruct the observed pattern using the correct counts per color. A pass requires an exact match of spatial configuration and color assignment. Task difficulty is defined on two axes: spatial load $L$ (number of occupied tiles) and feature-binding load $K$ (number of distinct colors), with feasibility constrained by $K \leq L$. Patterns are generated as single 8-connected clusters (i.e., all tiles in the cluster are contiguous along cardinal and/or intercardinal directions), are colored from a fixed, high-contrast palette and are generated such that difficulty is comparable for all equivalent $L$, $K$ stimuli.

Two administration modes were used. In Adaptive Mode (AM), participants completed a fixed sequence of 30 trials spanning the two-axis ($L$, $K$) domain. After every outcome (pass/fail), a Bernoulli-likelihood Gaussian Process (GP) classifier is updated online and the next ($L$, $K$) is selected where predictive entropy is maximal on a dense candidate grid. Recommendations are snapped to the integer lattice and respect the $K \leq L$ feasibility constraint. To avoid abrupt jumps in difficulty before the approximate thresholds are better known, proposed points are limited to at most +2 beyond the largest previously sampled value on each axis. In Classic Mode (CM), difficulty is controlled by a one-up/one-down staircase that varies only $L$ on a trial-by-trial basis while holding $K$ fixed at 3. The sequence starts at $L = 1$ and follows standard increment/decrement rules until the termination criterion is met.

## Platform and Administration

pixelDOPA (Digital Online Psychometric Assessment) is a custom-designed, unified, interactive, screen-based, 3D-rendered assessment environment. Details of how pixelDOPA is deployed as an experimental platform can be found at (Marticorena et al., 2025). Using a Dell Alienware m18 R2 laptop in the laboratory, participants connect to a secure server, enter a central lobby and launch the Build Master task. A brief standardized in-lobby tutorial introduces movement and interaction controls.

## Task

### Overview

As detailed in **Figure 1**, each trial comprises an acquisition (Observation) phase followed by a recall (Build) phase. In the Observation phase, a connected two-dimensional pattern occupies a fixed 5×5 grid. Initially seeing only gray tiles, participants must right-click on individual tiles to reveal their colors. When clicked, the tile's color remains visible for 1000 ms, after which time the entire tile disappears. Participants may reveal tiles in any order. The total duration of the observation phase is $5 \times L$ seconds or until all tiles have been revealed, whichever occurs first.

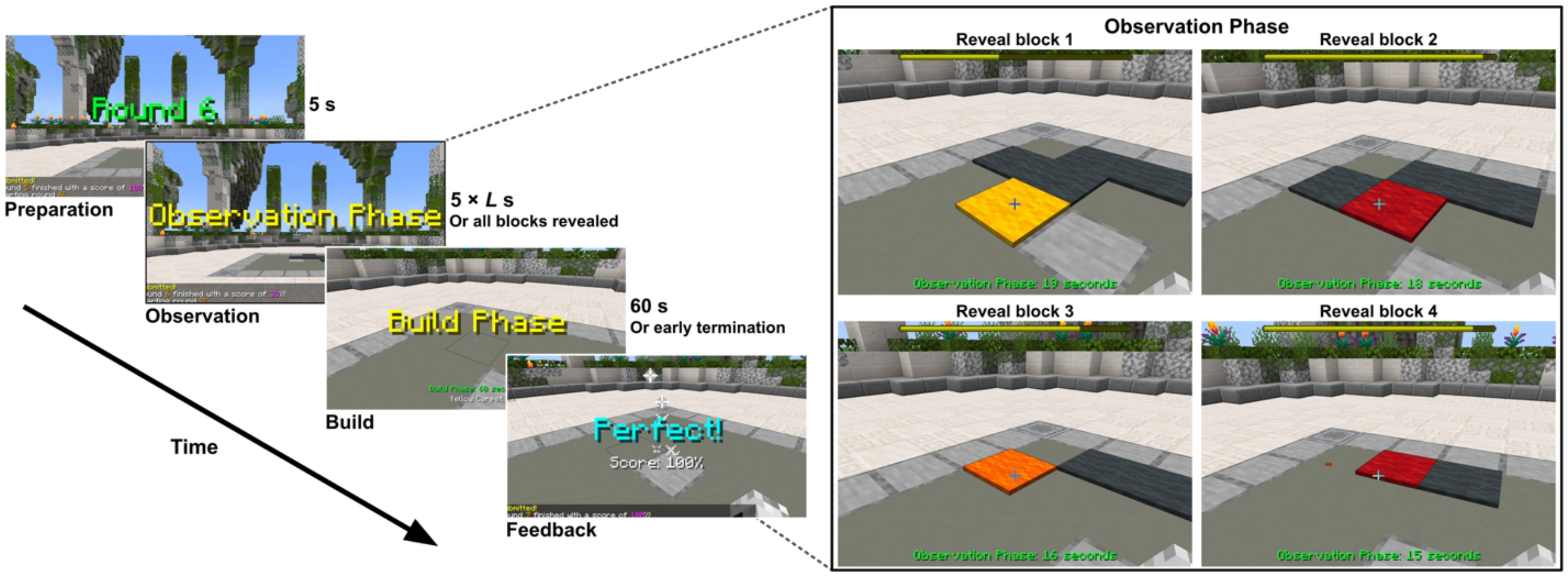

**Figure 1**. Task timeline and phases. (Left) Each trial proceeds as Preparation, Observation, Build, Feedback. The same temporal structure is used in Classic Mode (CM) and Adaptive Mode (AM). (Right) Screen captures show participant-initiated color reveal in the Observation phase for 4 blocks out of a 5-block sequence.

In the Build phase, participants are provided the exact count of tiles per color required to reconstruct the pattern presented in the Observation phase. Participants may place tiles onto the grid in any order and can remove and/or replace tiles after their initial placement. The phase ends after participants submit their pattern for scoring or the expiration of a 60-second timer, whichever occurs first. The recreated pattern is scored against the one presented in the Observation phase, with a pass requiring a 100% match of both spatial configuration and color assignment. Correct/Incorrect feedback and the recreation accuracy score are presented to participants briefly after each trial.

While the above description provides the actual experimental details, one of the great advantages of pixelDOPA is the ease with which experimental designs can be modified in advance and the potential for mining the detailed process data logged during every participant's testing to identify more informative data streams. Machine learning analysis of inhibitory control pixelDOPA process data, for example, has already revealed better indicators of participant cognition than the original design variables (Marticorena et al., 2025). Candidates for this type of *post hoc* evaluation in the current study include scoring partial matches for incorrect pattern builds and timing + movement data on how observation and building were accomplished. This capability showcases the potential for readily extending behavioral assessments in an immersive virtual environment, but the current study focuses exclusively on the conventional analytics described here.

## Stimulus Generation

For each (*L*, *K*) combination, candidate patterns were generated and scored before one representative pattern was selected for testing. Spatial patterns to be memorized were presented on a fixed 5×5 grid with exactly *L* occupied cells and *K* colors (with $K \le L$).

Stimuli were generated as single 8-connected clusters of size *L*, grown from a random seed cell with an 8-neighborhood frontier. Layouts with horizontal, vertical or diagonal symmetry were discarded. Each valid layout received a spatial-entropy score in [0,1] computed as a weighted blend of normalized mean pairwise Manhattan distance between occupied cells (favoring spread) and 1 minus the local 8-neighbor clustering coefficient (down-weighting tight clumps). We retained a large set of layouts per (*L*, *K)* to compute empirical percentiles.

For each retained layout and *K*, colors were distributed as evenly as possible across the *L* cells (counts differ by at most one) from a fixed, ordered palette of eight high-contrast hues (red, orange, yellow, lime, light blue, purple, pink, white), using only the first *K*. Many colorings were generated by randomly shuffling these assignments across cells occupied by tiles, including at least one deterministic coloring in which same-colored tiles are deliberately grouped into spatially contiguous clusters rather than randomly scattered. This process ensures the sample consistently spans the full range of dispersed to clustered color distributions. The spatial entropy score was computed as $S = 0.5 \times \tilde{D} + 0.5 \times (1 - \tilde{C})$, where $\tilde{D}$ is the normalized mean pairwise Manhattan distance between occupied cells and $\tilde{C}$ is the local 8-neighbor clustering coefficient. These fixed weights were design choices in the pattern generator rather than quantities fit from participant data. Each coloring was scored by a color-mix ratio = (adjacent different-color pairs) / (all occupied adjacencies) under 8-neighbor adjacency. A single pattern for a given ($L$, $K$) was then selected by minimizing $|P_S - 50| + |P_C - 50|$, where $P_S$ and $P_C$ are the empirical percentiles of the spatial-entropy and color-mix scores across all sampled ($L$, $K$) pairs. Generation used fixed pseudorandom seeds so identical inputs (($L$, $K$), target percentile pair ($P_S$, $P_C$), seed) reproduce the same pattern. Representative 1st, 50th, and 99th percentile examples are shown in **Figure 2**.

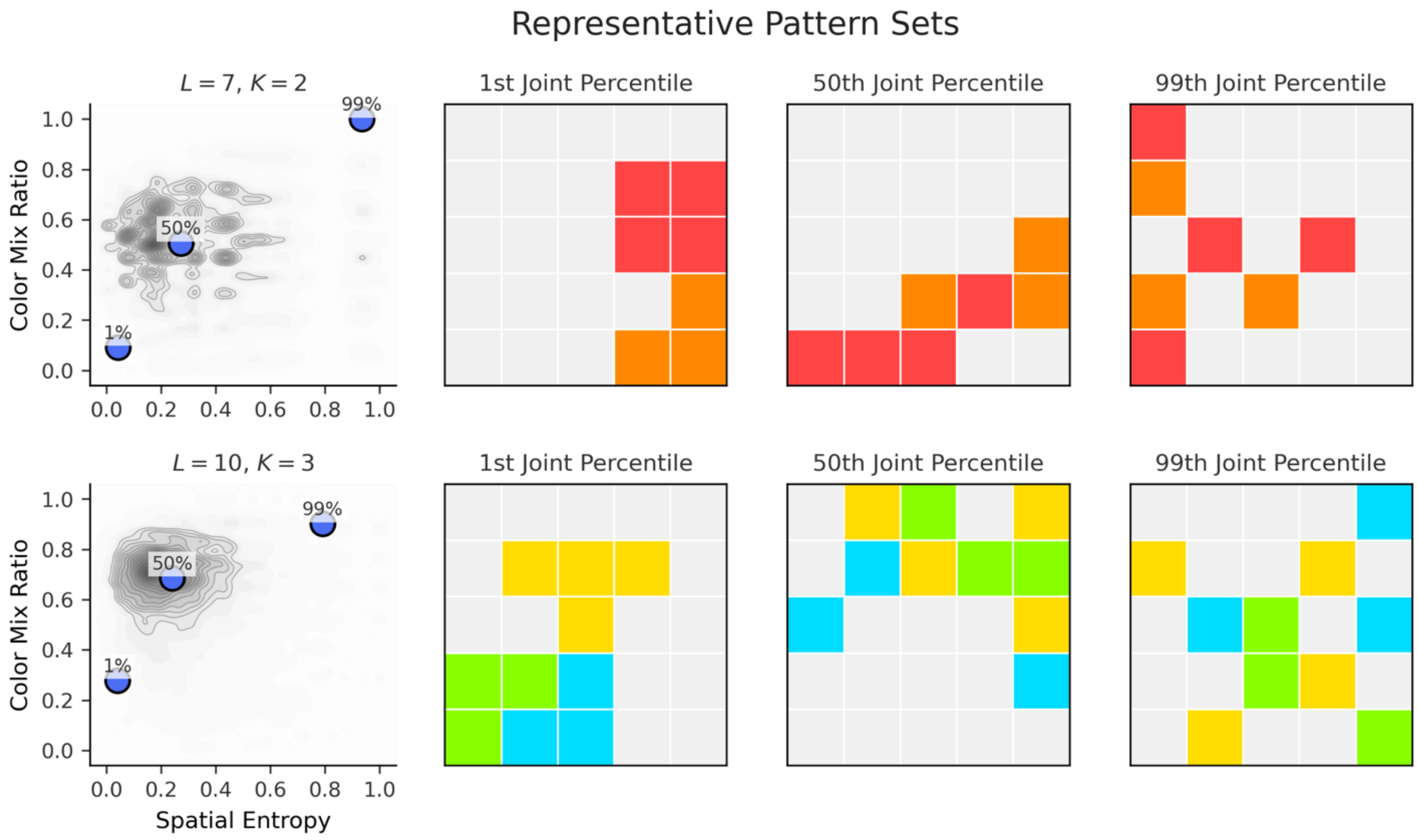

**Figure 2**. Stimulus standardization across the (*L*, *K*) domain. The top (*L* = 7, *K* = 2) and bottom (*L* = 10, *K* = 3) panels display the spatial entropy (abscissa) and color-mix ratio (ordinate) distributions on the left column, along with exemplar patterns at the 1st, 50th and 99th percentiles. All patterns used for this study were drawn from the 50th joint percentile set.

This procedural auto-generation of patterns ensures an extremely large number of possible stimuli that need not all be predetermined by hand, as they must be in Corsi tasks (Kessels et al., 2000), for example. Selecting the 50th percentile for all patterns ensures that the nature of the patterns themselves does not contribute to task difficulty independent of *L* and *K*, thereby reducing the model complexity (i.e., 2D vs 3D) required for the estimation computation. Should it prove interesting or useful to incorporate pattern entropy as another independent variable into the estimation problem, it is straightforward to include it in the GP model as an additional variable.

Models up to 10th order have been successfully evaluated using similar active learning procedures (Heisey, 2020).

## Experimental Modes

Two separate modes of the Build Master task were administered: Adaptive Mode (AM), and Classic Mode (CM).

### *Adaptive Mode*

Patterns are parameterized by $L \in [1, 16]$ and $K \in [1, 8]$. The first two trials for each testing session are non-adaptive primer trials with ($L$, $K$) = (1, 1) and (3, 3). To ensure trials maintain a smooth progression and avoid large increases in difficulty and cognitive load, all proposed points selected by the optimization algorithm are capped at +2 beyond the largest previously sampled value on either axis. Selected points can freely become less difficult, however. This “dampening” approach is mimicked in other entropy-maximization acquisition functions wherein subject comfort is taken into account, such as hearing tests (Song et al., 2015). Mathematical details of the GP implementation can be found at (Marticorena, Wong, Browning, Wilbur, Davey, et al., 2024).

### *Classic Mode*

In CM, difficulty follows a one-up/one-down staircase over pattern length. We fix the number of colors at $K = 3$ and vary the number of tiles $L$ by ±1 between consecutive trials, incrementing $L$ after a correct recollection (i.e., “pass”) and decrementing after an incorrect one (i.e., “fail”). The sequence initializes at $L = 1$ and proceeds until termination criteria are met: either two failures occur at the same $L$, or the participant successfully recreates the maximum size ($L = 16$).

This implementation is meant to make the active-learning procedure transferable across behavioral tasks rather than specific to Build Master. For a new task, the experimenter defines the stimulus dimensions to be modeled, their allowable ranges and any feasibility constraints on candidate stimuli. The task client then returns an outcome after each trial, typically a binary pass/fail label or a threshold-relevant transformation of a graded response. Optional phantom observations can be added when boundary behavior is known in advance, such as likely success in trivially easy regions or likely failure beyond the practical task range. Once these task-specific elements are specified, the remaining operations are handled by the same API workflow used here: inputs are scaled internally, the GP is updated after each labeled observation, posterior uncertainty is evaluated over the feasible candidate set and the next recommended stimulus is returned in the task’s native coordinates. The same serialized outputs, including sampled points, masks, posterior grids and learned hyperparameters, can then be used to extract thresholds, isocontours or other task-relevant summaries.

## Adaptive Policy and API Integration

We built an instrument-agnostic, real-time Application Programming Interface (API) service that selects the next ($L, K$) pair by updating a GP classifier after every trial and sampling where the predicted outcome is most uncertain. The service enables cloud-based active learning (Barbour et al., 2019) and follows with identical choices for GP model, link/kernel selection, variational training and optimization (Marticorena, Wong, Browning, Wilbur, Davey, et al., 2024). The Build Master client sends parameters (*L, K*) in native units, receives the service's next recommendation and returns a binary outcome (pass/fail). Primer trials from the instrument initialize a session and optional "phantom" observations. These boundary phantoms prevent the acquisition function from repeatedly targeting the feasibility boundary (L ≤ K), where posterior uncertainty is high by construction rather than informative about task performance. A sensitivity analysis removing all phantom observations confirmed that threshold estimates shift substantially and lose convergent validity with the classical staircase (Supplemental Section 5). Inputs are internally scaled to [0, 1] on both axes for numerical stability, but all reports use native units. After each outcome, the GP is updated online. After each new observation, the GP is updated incrementally while preserving the previous variational state. If this incremental update fails to converge, the model is retrained from scratch. The acquisition function evaluates predictive entropy on a dense grid and proposes the maximum-entropy candidate, with ties being broken by preferring points farthest from previously observed (non-phantom) samples to reduce clustering. The service respects user-specified feasibility constraints, such as integer-domain snapping, polygonal inclusion regions or step caps. In Build Master we enforce *K* ≤ *L* feasibility via a polygonal mask and snap both axes to integers. Analytical reproducibility is ensured via fixed seeds, thread-safe updates and automatic serialization of metadata, sampled points, masks, posterior grids and learned hyperparameters.

## Participants & Procedure

### Participants

Eligibility criteria include adults aged 18–30 with normal or corrected-to-normal vision, fluent English (native or demonstrably fluent) and basic computer interface familiarity. Prior general electronic gameplay or specific experience with the software underlying the pixelDOPA platform was not required or selected for. All procedures were approved by the Institutional Review Board at Washington University in St. Louis (protocol #202508115), and the experiment was explained in detail to participants.

Thirty-two adults enrolled; one was deemed ineligible based on English fluency. Two sessions were not captured because of live database failures, leaving behavioral data for 29 unique participants (mean age ± standard deviation = 21.0 ± 3.5 years, range 18–29). Eight participants completed a second session, and after exclusions (2 extreme outliers, 2 non-finite thresholds at $K = 3$) the analyzed dataset contained 33 sessions from 27 individuals, with six participants contributing a repeat session. Sex/gender among the analyzed participants: 17 male, 9 female, 1 other. Unless noted, inferential analyses are per session, and demographics are per participant.

### Surveys

Before gameplay, participants completed a brief survey capturing mouse/keyboard comfort, gaming background and familiarity with the software architecture for the pixelDOPA platform. After both modes were completed, a post-gameplay survey assessed focus, enjoyment for each mode, open-ended approach descriptions (including differences between modes) and mental fatigue after completing each mode.

## Procedure

Task-mode order was counterbalanced using an even/odd assignment: even-numbered participants completed CM followed by AM while odd-numbered participants completed AM followed by CM. Each session followed a standardized timeline within pixelDOPA: participants joined the server, completed a short in-lobby movement/interaction tutorial, and adjusted input settings (e.g., mouse sensitivity, key binds) as needed. The experimenter then briefed participants on the default metatask structure (Mode 1 → optional break → Mode 2). Within the Build Master task, a concise tutorial ensured comprehension, and up to five minutes of optional practice was available based on participant preference and/or experimenter judgement. Participants then completed Mode 1 (either AM or CM), had the option for an up to 5-minute break, then completed Mode 2 (the remaining mode), after which the post-gameplay survey was administered.

## Repeat Session

Participants were invited to return on another day to complete a repeat session. Those who returned performed the two modes in the opposite order from their first session (within-subject counterbalancing). The pre-gameplay survey was omitted and tutorial/practice elements were abbreviated based on prior familiarity. The post-gameplay session was readministered following the completion of both modes.

# Data Analysis

Performance in both modes is summarized by $\psi_\theta$, the 50% point of the psychometric link function. In AM, $\psi_\theta$ is read directly from the GP's predictive posterior at $K$ = 3. In CM, $\psi_\theta$ is obtained by fitting a logistic function to the staircase series. When applied to Corsi testing data in a similar population with similar numbers of trials, the CM procedure generates threshold estimates correlated at 0.9 with a common “max sequence length” test measure, implying that it is a reasonable metric to summarize spatial working memory (Rojo et al., 2023). Analyses draw on two sources: CM psychometric fits and AM GP posteriors. To ensure comparability, AM posteriors are standardized with a single refit, and a shared validity mask and outlier rule are applied once upstream, ensuring the same set is used throughout.

For AM posterior standardization, we recompute each session’s GP using only observed samples and a fixed boundary set of “phantom” observations. When a particular color $K^*$ has more passes than fails, we insert data-driven monotonicity phantoms (positive labels for all $L \leq K^*$). All labels are binary, and coordinates remain within task bounds. The 50% performance isocontour $\psi_\theta(K)$ is extracted as the first crossing of 0.5 along $L$ at each fixed $K$, using linear interpolation.

Thresholds are then derived per-modality. For CM, a logistic function is fit per entity and $\psi_\theta$ taken from the 50% probability of success. To guard against runaway estimates near ceiling, $\psi_\theta$ is capped at the task maximum and a single phantom failure is placed just beyond range to regularize the fit. For AM, $\psi_\theta$ is read directly from the predictive posterior. Validity filtering retains only paired thresholds wherein the estimated threshold for pattern length $L$ at $K = 3$ was above 0 for both adaptive and classic modes at $K = 3$. Outliers were defined on AM-CM differences in $\psi_\theta$ using a 1.5 interquartile-range fence. The agreement between AM and CM thresholds was quantified with the two-way random effects, absolute-agreement, single-measure ICC (2,1). Statistical evidence relied on the associated *F*-test (between- vs. within-session variance) and 95% confidence intervals obtained from the *F*-distribution bounds on that ratio (Motsnyi, 2018). Correlational summaries paired CM $\psi_\theta$ with survey covariates (pixelDOPA software architecture familiarity, mouse/keyboard comfort, etc.) using Pearson's *R*, evaluated by the standard two-tailed *t*-test (df = $n - 2$) with Fisher *z*-based 95% CIs. Each correlation also reports its coefficient of determination ($R^2$) and a Bayes factor $BF_{10}$ computed under the Jeffreys-Zellner-Siow (JZS) prior (Bayarri & García-Donato, 2008). Session-order effects were analyzed within each order group by forming the paired difference Δ = Session 2 – Session 1 and applying a two-sided, one-sample *t*-test on Δ, alongside its Student-*t* 95% CI and matching $BF_{10}$. For the difference test between Independent Staircase and Active sampling, we used the above paired differences *t*-test, with equivalence at the earliest sample using the above JZS $BF_{10} \leq 3$.

## Simulated Experiments

To evaluate the relative performance of different trial-selection (i.e., sampling) schemes, the ability of these machine learning models to generatively produce item-level data is exploited to simulate virtual participant sessions matched in performance to the sessions of real participants. Each human session's standardized AM $\psi_\theta(K)$ is converted into a cumulative normal distribution function generator over the task domain with individualized guess and lapse; $\psi_\theta(K)$ is smoothed across *K* with cubic splines, and spread is bounded away from zero for stability. These known ground-truths are used to make absolute determinations of algorithm performance.

Three sampling procedures are compared: Independent Staircase (one-up/one-down at $K$ = 1…8, cycling with the CM termination logic), Halton low-discrepancy sampling and Active entropy maximization (API driven). Halton sets are deterministic space-filling sets that sample the full range of input variable values for multidimensional models (Song, Garnett, et al., 2017; Song, Sukesan, et al., 2017). Because Halton sampling does not use response feedback, it serves as a deterministic space-filling control that helps separate the benefit of broad space coverage from the benefit of entropy-guided adaptive selection. All runs start with two identical primer trials, ($L$, $K$) = (1, 1) and (3, 3), then proceed sample-by-sample to 100 total. After each observation the GP is refit online and $\psi_\theta(K)$ is re-extracted. Accuracy at step $t$ is the Root Mean Square Error (RMSE) between the estimated and ground truth 50% isocontours. For fixed budgets (e.g., 30 samples and final), one-shot posteriors are computed on cumulative samples (plus the same monotonicity phantoms) to enable fair overlays. Uncertainty bands around isocontours are derived from posterior percentiles (30–70%). Session trajectories are aggregated as mean and dispersion across the samples.

# Results

## Experiment 1

We first validated the two-axis Gaussian Process (GP) active classifier against a conventional one-axis staircase. Classic Mode (CM) delivered a mean of 14.1 trials per session (SD = 4.17; range = 9–26). Adaptive Mode (AM) used a fixed 30-trial budget across the full ($L$, $K$) domain; of these, the model delivered a mean of 5.48 trials to $K$ = 3 (SD = 1.77; range = 2–10).

**Figure 3** reveals session heterogeneity that a single-axis procedure cannot expose. Sessions with similar CM $\psi_\theta$ often exhibit very different two-dimensional posteriors: some isocontours decline steeply with $K$ (i.e., strong $L{\times}K$ trade-off) while others are comparatively flat (i.e., robustness to added color bindings). Several isocontours show local curvature or coverage differences along $K$ that CM could not reveal. Visual inspection shows isocontour slopes ranging from −2.31 to −0.71 $\Delta K/\Delta L$ for sessions near CM $\psi_\theta = 7.5$.

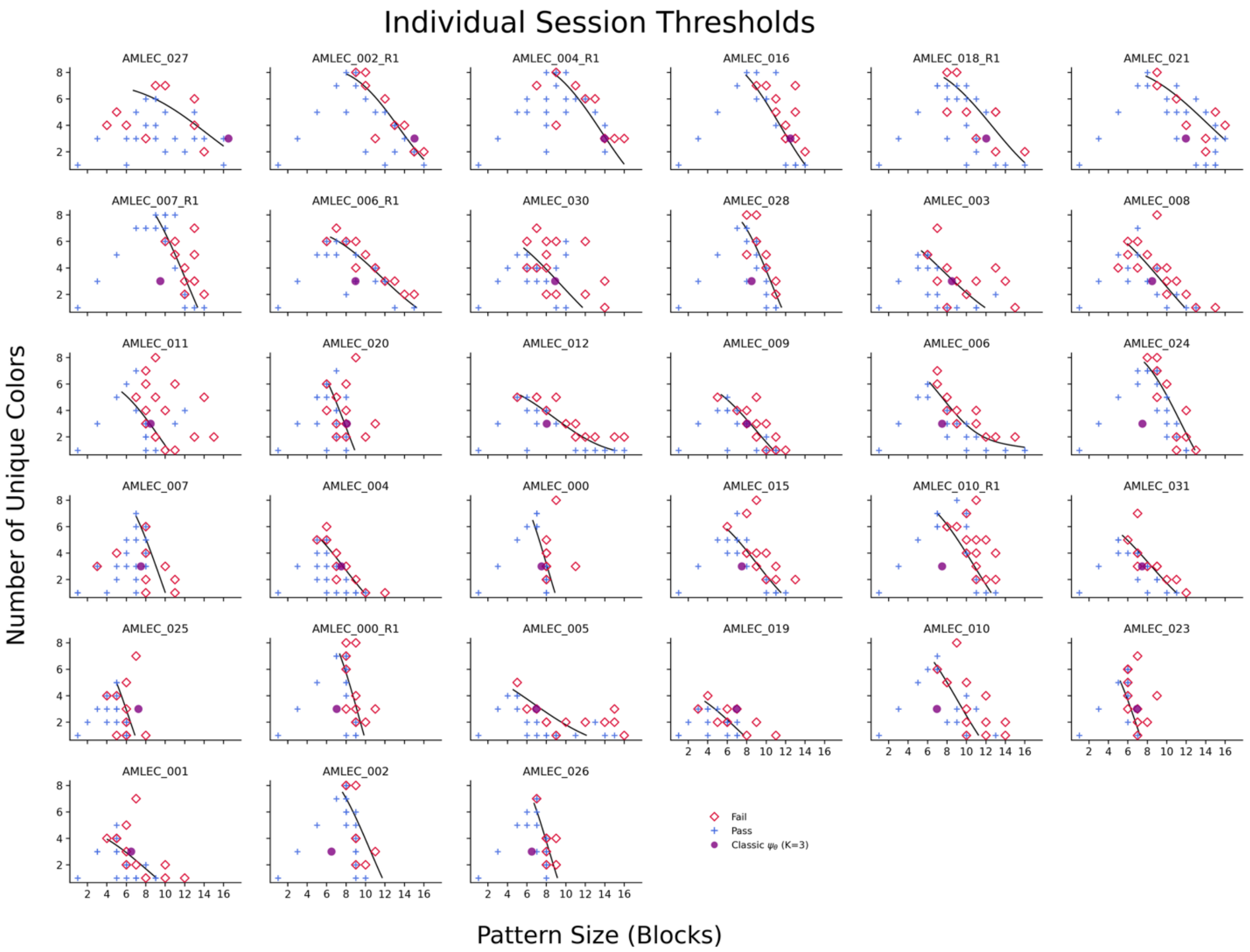

**Figure 3.** Two-dimensional performance domains for individual sessions sorted by descending CM $\psi_\theta$. Plotted are AM samples (pass = blue crosses, fail = red diamonds), the GP-derived 50% performance isocontour (black lines) and the CM $\psi_\theta$ at $K$ = 3 (purple dots).

Agreement between CM thresholds and AM thresholds at $K = 3$ is strong (**Figure 4**). The single-measure two-way mixed-effects ICC(2,1) was 0.755 with 95% CI [0.574, 0.878], $p = 3.96\times10^{-9}$, $BF_{10} = 2.65\times10^{6}$ ($n = 33$). We chose ICC over Pearson's *r* because ICC(2,1) penalizes both systematic bias and random disagreement, whereas Pearson's *r* is invariant to constant offsets. This distinction is important here because AM thresholds were modestly but systematically higher than CM thresholds, despite strong rank-order agreement. Supplemental trial-matching analyses showed that unequal trial counts attenuated but did not eliminate this offset, and locality-matched GP analyses showed that the offset was reduced when the GP estimate was forced to rely primarily on observations near $K = 3$. These results indicate that cross-*K* information borrowing contributes to the AM–CM difference, while preserving the main conclusion that a two-axis GP classifier recovers broadly comparable $K = 3$ thresholds while simultaneously learning across the full ($L$, $K$) domain.

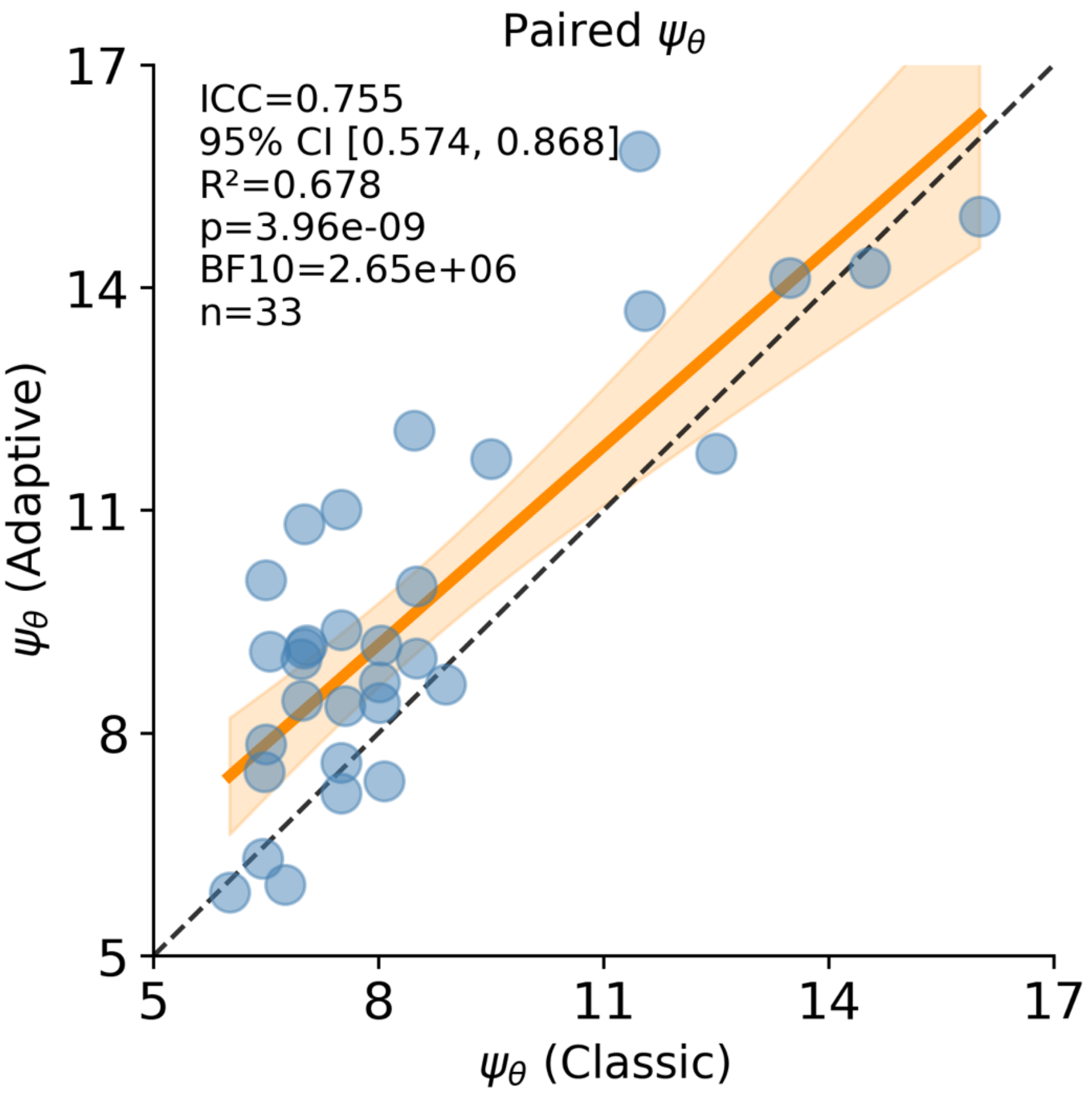

**Figure 4.** Agreement between Adaptive and Classic $\psi_\theta$ at $K$ = 3. Each point is one session. The dashed line marks identity; the solid orange line is the linear fit with 95% CI. Values printed on the panel give agreement statistics (ICC(2,1), $p$ value, $BF_{10}$), coefficient of determination $R^2$, and sample size. $\psi_\theta$ is the 50% point of a fitted sigmoid for CM and the 50% point of the GP classifier's predictive posterior for AM.

The distribution of within-person changes (Session 2 – Session 1) for sessions starting with CM was centered well above zero (mean $\Delta\psi_\theta$ = +1.62, $n$ = 14, $p = 9.83\times10^{-4}$), as visible in the summary of **Figure 5**. In contrast, sessions starting with AM exhibited little difference between Modes 1 and 2 (mean $\Delta\psi_\theta$ = −0.650, $n$ = 19, $p$ = 0.114), with the bulk of the distribution straddling zero. This disparity may reveal additional nuance from optimal adaptive WM assessment compared to conventional adaptive staircases.

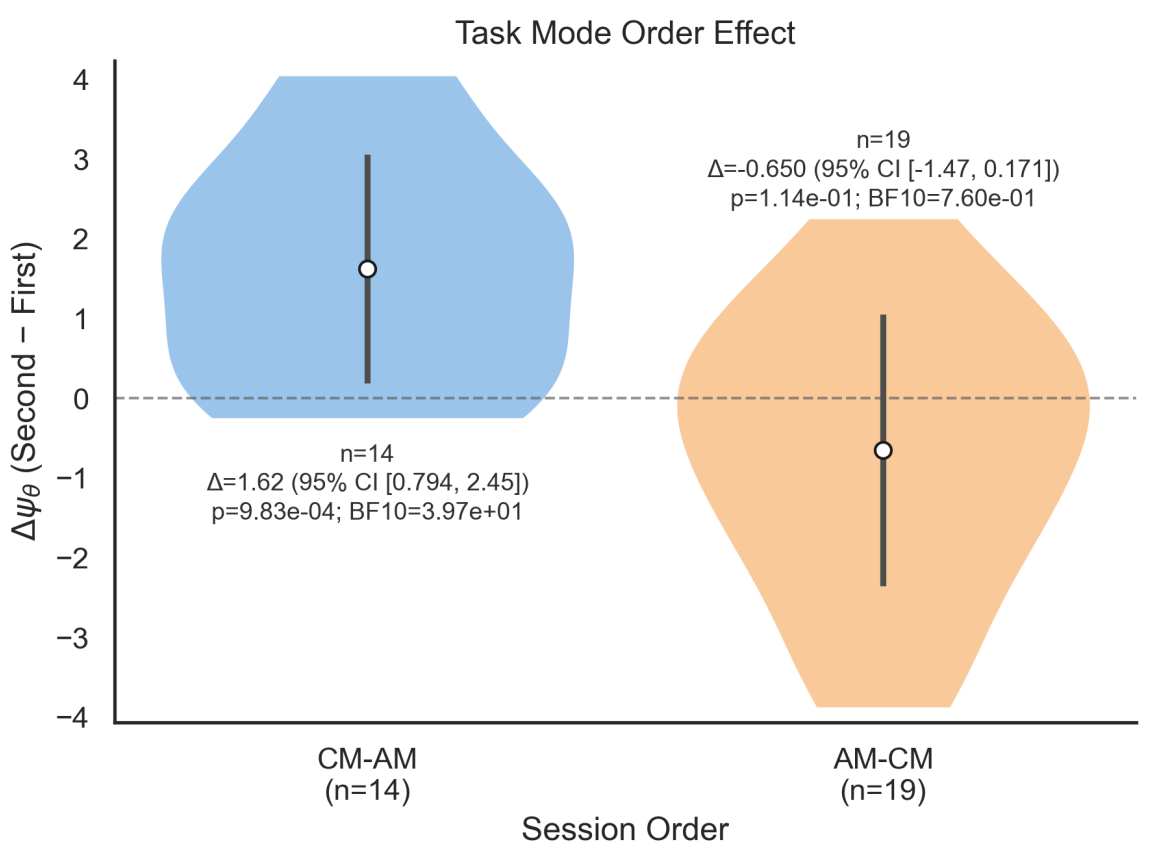

**Figure 5.** Session-order effect on $\psi_\theta$. The y axis shows (Session 2 $\psi_\theta$ − Session 1 $\psi_\theta$), split by mode order (CM followed by AM, AM followed by CM). Medians and *p*-values are given beneath each distribution. Sessions that began with CM showed significant improvement on their second mode (AM) (median +1.73), whereas the opposite order revealed no significant change.

Self-reported familiarity with the software architecture underlying the pixelDOPA platform showed no meaningful relationship to classic $\psi_\theta$: the correlation was undetectable ($r$ = 0.200, $p$ = 0.339, 95% CI [−0.212, 0.551], $BF_{10}$ = 0.458).

Mouse/keyboard comfort showed a similar, but opposite trend ($r$ = −0.176, $p$ = 0.380, 95% CI [−0.521, 0.219], $BF_{10}$ = 0.414). Therefore, little evidence exists at this point for a relationship between familiarity and task performance.

## Experiment 2

Simulated runs of generative models allowed us to test sampling efficiency and precision. **Figure 6** plots the RMSE versus sample count for the Independent Staircase (1-up/1-down), Halton and Active Acquisition.

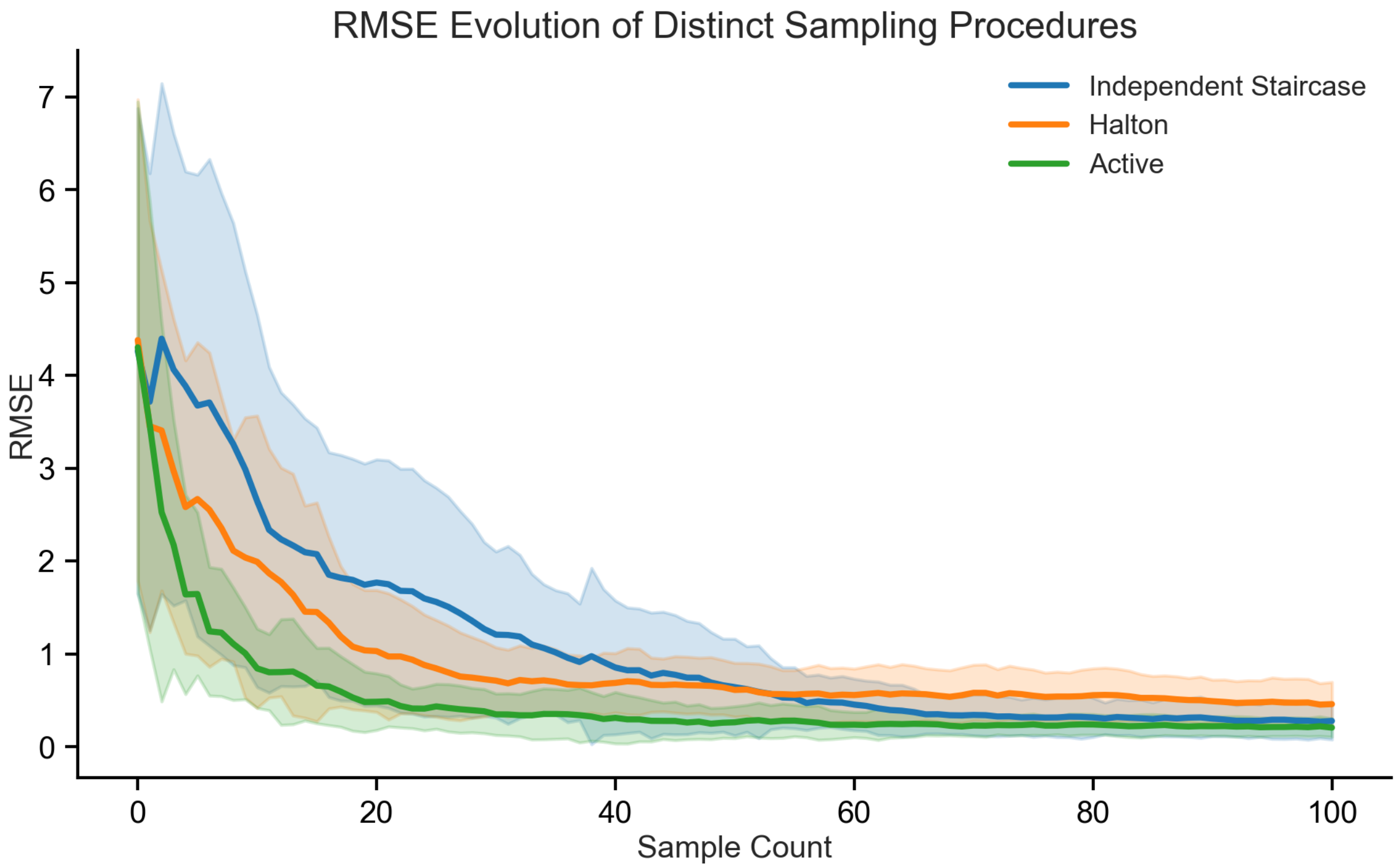

**Figure 6**. Root Mean Square Error (RMSE) versus sample count for simulated data. Using 33 session-specific generators, we simulated three sampling policies after a two-trial primer: Independent Staircase (blue), Halton (orange) and Active entropy-maximization (green). Curves show mean RMSE ±1 SD across 33 sessions as the GP is re-fit after each additional observation (up to 100).

Active sampling reaches low error rapidly and maintains the lowest RMSE and tightest spread across sessions across the entire 100-sample horizon (Figure 6). Independent Staircases converge slowly and noisily, with large between-session variability at the practical budgets used in Experiment 1 (about 25–30 samples). Halton improves upon staircases but trails Active on both accuracy and precision. By roughly 25 samples, Active has largely stabilized, empirically justifying the 30-trial AM budget. At 30 samples, Independent Staircase has a measurably elevated mean RMSE compared to Active with a $\Delta\mu_{RMSE}$ of 0.887, 95% CI [0.556, 1.22], $p = 5.12\times10^{-6}$, $BF_{10} = 3.85\times10^{3}$ ($n = 33$, Cohen's dz = 0.951). With enough samples the two methods equilibrate in $\mu_{RMSE}$ in that $BF_{10} \leq 3$ only at 66 samples and beyond.

**Figure 7** illustrates predictive posterior cross-sections at a fixed budget of 30 samples. The dotted line is the ground-truth 50% isocontour, solid lines are each method's estimate, and shaded regions show ±20% posterior bands. Most sessions show Active nearly overlapping the ground truth with narrow bands, while Independent Staircases deviate and show inflated uncertainty at low *K* or high *L*. In a few challenging cases for Active (e.g., 006 and 007-R1), the fit slightly under- or over-shoots a plateaued ground-truth segment. Yet even there, its uncertainty remains smaller than Halton's and far smaller than Independent Staircases'. The mean fit remains closer to ground truth, as well.

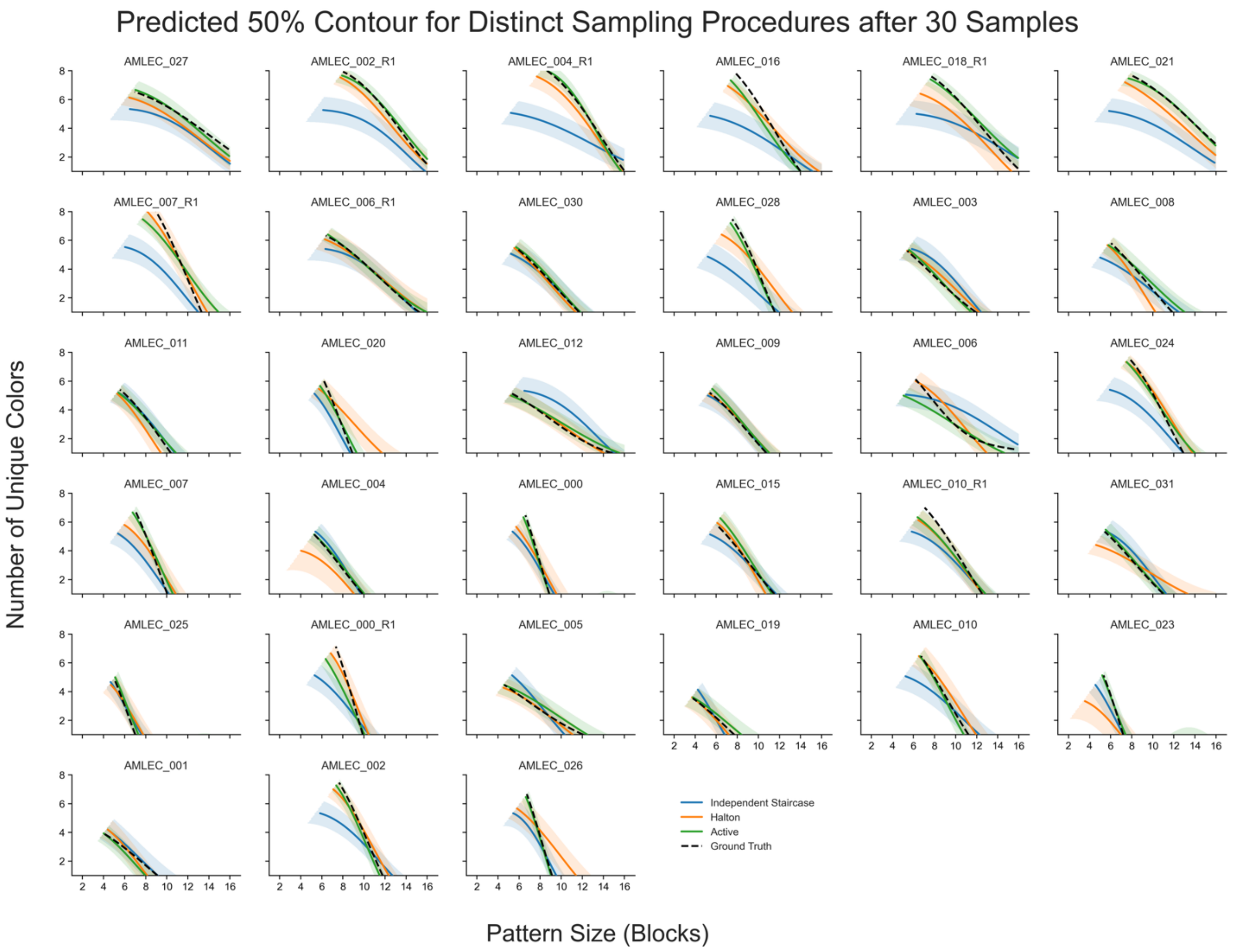

**Figure 7**. Contour quality at a fixed budget (30 samples) for simulated sessions in the same sort order as Figure 3. For each session, the dashed curve is the ground-truth 50% isocontour; solid curves are the estimated 50% isocontours from Independent Staircase (blue), Halton (orange) and Active (green). Shaded regions depict each method's ±20% posterior band (0.3–0.7).

# Discussion

In a population of healthy young adults, our multidimensional, GP-based active machine learning procedure recovers estimates of spatial WM thresholds highly comparable to a conventional one-dimensional staircase, while simultaneously learning a richer picture of performance across stimulus factors. Despite fewer $K = 3$ samples, AM's $\psi_\theta$ still agrees strongly with CM, and the simulations in Experiment 2 clarify that entropy-guided sampling concentrates information and recovers accurate isocontours with far fewer samples per slice.

AM provides access to the shape of the working memory boundary, beyond set size, length or span. The plots in Figure 3 make this finding clear: session results that appear similar by a single value (i.e., their CM $\psi_\theta$) often differ substantially in two dimensions. Some isocontours fall sharply as $K$ increases, revealing a strong load-binding trade-off, while others are comparatively flat, indicating tolerance to additional color bindings at a fixed spatial load. Several isocontours bend or plateau toward $K = 1$ pointing to potential non-linear $L \times K$ interactions. While this exploratory methodological study did not set out to test hypotheses about the interactions between spatial working memory and feature binding, this finding does confirm that WM estimation using a multidimensional GP with active learning can resolve such patterns efficiently and opens the door for a wide variety of study designs.

Although our primary aim was methodological, the individual variation in isocontour geometry is itself informative. Across participants, the 50% contour varied substantially in slope, indicating that increases in color load did not reduce tolerable spatial load by a fixed amount. Some participants showed steep negative contours, whereas others showed much flatter profiles. This heterogeneity is invisible to any single-axis staircase and suggests that individuals differ not only in overall capacity but also in how spatial and feature-binding demands interact, a distinction relevant to both slot-based and resource-based accounts of working memory. Future work could test whether isocontour shape predicts transfer, training responsiveness, or clinical status.

Machine learning models are generally quite flexible and typically require large amounts of data to fit them. We pioneered the use of GPs as probabilistic classifiers to generalize the psychometric function to multiple dimensions with the potential to quantify nonlinear interactions (Song et al., 2015). Adding modern optimizers for making optimal trial selections (Kingma & Ba, 2017) and incorporating Bayesian priors enables these models to be fit with sparse data, as we have demonstrated in multiple behavioral domains. The resulting models are efficient and informative because embeddings or projections take on interpretable forms.

In the current study, the GP classifier's sigmoidal link function lets us read $\psi_\theta$ anywhere in the stimulus domain while retaining a calibrated notion of uncertainty everywhere else. Coupling that model with entropy-based acquisition rapidly concentrates trials near informative regions, thereby reducing participant fatigue without sacrificing inferential power. That combination is especially attractive for WM tasks because each one-factor staircase is inefficient: mapping a two-factor domain would require many one-factor staircases and thus invite fatigue. Furthermore, because the $L \times K$ relationship differs across individuals, a $\psi_\theta$ measured at a single $K$ does not generalize, masks $L \times K$ interactions and understates uncertainty (Figures 3, 6 and 7). By contrast, AM estimates the full predictive posterior surface and achieves accurate isocontours within roughly 25–30 trials.

Experiment 2 grounds these results in study design terms. Using participant-session-specific generative models, we quantified accuracy as a function of sample count and allocation policy. Active sampling reached low error rapidly and stabilized by ~25 samples, with consistently narrower between-session dispersion than Independent Staircases or Halton (Figure 6). At a budget that matches our real session (30 trials), AM not only tracked the ground-truth isocontours more closely on average; it also delivered tighter posterior confidence bands (Figure 7). The panel-by-panel inspections are instructive: in most cases, the green isocontour largely coincides with the dashed ground truth, while in other cases (e.g., 006, 007-R1) Active sampling occasionally under- or overshoots a local plateau, but still with markedly less uncertainty than the alternatives. These results argue for information-optimal allocation when the aim is to recover a multidimensional threshold.

We observed a clear order effect given that $\psi_\theta$ was significantly larger when CM preceded AM, but the difference was near zero when AM preceded CM (Figure 5). Because mode order was counterbalanced, this finding does not threaten the main inferences but does add noise to the observed trends. It also raises interesting questions about why it occurred. Because some individuals seem to be able to self-identify learning strategies upon repeated sessions of WM training (Feng et al., 2023), it is possible that the AM sampling approach of presenting spatial patterns for recall in no clear order from the participant's perspective prompted more exploratory encoding approaches in participants than the easily predictable staircases of CM. Because variable sampling approaches represent information masking for perceptual modeling, the issue of threshold bias has been evaluated from the first use of active GPs to model behavior without finding an effect (Song et al., 2015). The effect of this phenomenon on cognitive assessment and training studies is worthy of further investigation.

We note three limitations that qualify the overall study findings and point to clear next steps. First, parity was established at a single comparison slice ($K = 3$). Time and fatigue limits prevented us from running dedicated staircases at other $K$ values. Future work could validate parity across multiple slices or use brief “check-staircases” at strategically chosen $K$ to audit the AM domain. The supplemental locality-matched analysis further suggests that validation at additional $K$ values would help separate true cross-dimensional generalization from GP smoothing effects near a single comparison slice. Second, the simulation ground truths were derived from each human session’s fitted GP. This choice is correct for isolating sampling policy (all estimators used the same modeling family), but it also makes the comparison conservative for the GP-based Active method. Cross-family comparisons (e.g., parametric QUEST+ variants or hierarchical linking across participants) would broaden the scope. Third, the slope analysis is underpowered and should be treated as hypothesis-generating. With a larger sample, we can test whether isocontour geometry (slope, curvature or area under the feasible region) predicts transfer, training responsiveness or something else of interest as a candidate biomarker rather than a mere description.

Practically, Adaptive Mode serves two use cases. AM delivers reasonable estimates of $\psi_\theta$ at a single $K$ condition while also fitting the full ($L, K$) domain. If investigating how two determinants of memory performance jointly shape behavior (e.g., spatial load with binding demand, delay with interference, set size with distractor similarity), then one-factor staircases are inadequate, and AM provides the needed isocontour and its uncertainty. Our approach is particularly valuable when the relationship between the stimulus dimensions is unknown or not well-fit to a parametric function. In our data, about 30 active trials typically yield a stable 50% isocontour. When finer precision is needed, it is straightforward to add active queries targeted to the most uncertain segments of that isocontour. Because the adaptive policy is instrument-agnostic, the same procedure should transfer to other working-memory paradigms (e.g., color-shape binding sequences) and to settings where multidimensional difficulty or stimulus delivery is the norm. An application of active machine learning for real-time multidimensional executive function testing is provided in our companion article in this issue.

# Conclusion

We validated a multidimensional Bayesian active classification approach to working-memory assessment that learns a participant-specific performance surface over spatial load $L$ and feature-binding load $K$. With only ~30 actively sampled trials, a 2D Bernoulli-likelihood Gaussian process classifier recovers 50% performance isocontours and yields $\psi$-equivalent thresholds at $K = 3$ that agree closely with a conventional one-up/one-down staircase, while simultaneously revealing exploratory individual variation in $L \times K$ geometry that the staircase method cannot capture. Simulations grounded in participant posteriors show that entropy-guided active sampling concentrates measurements near informative regions and reaches accurate isocontours with markedly fewer observations than baseline policies. Practically, this procedure enables short sessions that return both a parity point and a calibrated 2D difficulty map. Because the adaptive policy is instrument-agnostic and constraint-aware, the procedure should transfer to other multidimensional paradigms. More broadly, combining nonparametric probabilistic modeling with information-optimal selection offers a general recipe for scalable, precise psychometric estimation in complex task domains within experimentally manipulable immersive environments.

# Declarations

## Funding

The research reported here was supported by T32NS115672 and grants from the Washington University Office of the Vice Chancellor for Research and Here and Next Program.

## Conflicts of Interest/Competing Interests

The authors declare that they have no conflicts of interest or competing interests.

## Ethics Approval

All procedures performed in this study were approved by the Institutional Review Board at the Washington University in St. Louis (202508115) and adhered to the ethical standards of the 1964 Declaration of Helsinki and its later amendments or comparable ethical standards.

## Consent to Participate

Written study information was provided to all participants prior to their inclusion in the study.

## Consent for Publication

Not applicable.

## Availability of Data and Materials

Project information, including the data necessary to replicate these analyses and supplemental material, can be found at https://osf.io/ukbpn/.

## Code Availability

The code necessary to replicate these analyses can be referenced at https://osf.io/ukbpn/. The public API exposes a simple session workflow: initialize a session with labeled two-dimensional points and axis bounds, iteratively send binary outcomes back to the model, and query the next recommended point. Because the API preserves the user's coordinate system and supports optional inclusion polygons and discrete sampling constraints, it can be adapted to a broad range of two-dimensional psychometric tasks. Full documentation, environment files, and worked examples are included in the repository.

## Authors' Contributions

Dom CP Marticorena: Experimental design, experimental data collection, data analysis, manuscript drafting.

Chris Wissmann: Experimental design, experimental data collection, data analysis, manuscript drafting.

Zeyu Lu: Data analysis, manuscript editing.

Dennis L. Barbour: Overall supervision, experimental design, funding acquisition, manuscript finalization.

All authors read and approved the final manuscript.

## Open Practices

The authors followed guidelines recommended by the American Psychological Association (APA) for reporting methods and analyses.

# Multidimensional Bayesian Active Machine Learning of Working Memory Task Performance

Dom CP Marticorena[1], Chris Wissmann[1], Zeyu Lu[2], Dennis L Barbour[1,2]

[1]Department of Biomedical Engineering, Washington University, 1 Brookings Drive, St. Louis, MO 63130
[2]Department of Computer Science and Engineering, Washington University, 1 Brookings Drive, St. Louis, MO 63130

## Supplemental information for the manuscript with the same title

# 1 Extra Information

**Table S1.1**: Demographic details of study participants.

| category | level | n | percent |
|---|---|---|---|
| **Sex/Gender** | Male | 19 | 65.52 |
| **Sex/Gender** | Female | 9 | 31.03 |
| **Sex/Gender** | Other | 1 | 3.45 |
| **Hispanic/Latino** | None of these | 28 | 96.55 |
| **Hispanic/Latino** | Yes | 1 | 3.45 |
| **Race** | Asian | 15 | 51.72 |
| **Race** | White | 10 | 34.48 |
| **Race** | Black or African American | 2 | 6.9 |
| **Race** | White,Black or African American | 1 | 3.45 |
| **Race** | White,Asian | 1 | 3.45 |
| **Education** | Some college but no degree | 12 | 41.38 |
| **Education** | Bachelor's degree in college (4-year) | 8 | 27.59 |
| **Education** | High school graduate (high school diploma or equivalent including GED) | 5 | 17.24 |
| **Education** | Master's degree | 4 | 13.79 |

# 2 Alternate & Extra Analyses of Data Included in Manuscript

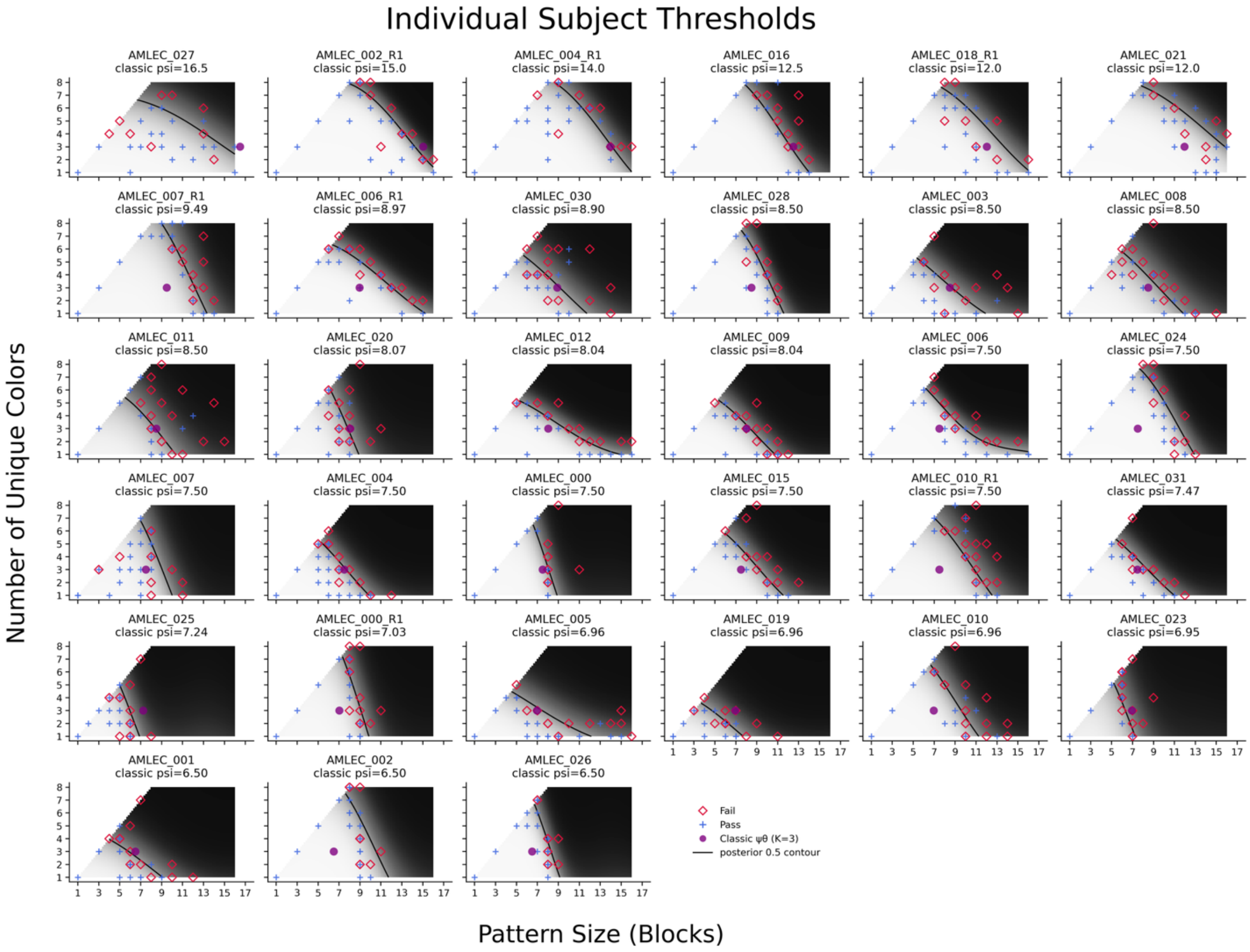

**Figure S2.1**: Two-dimensional performance domains for individual sessions sorted by descending CM $\psi_\theta$. Plotted are AM samples (pass = blue crosses, fail = red diamonds), the GP-derived predictive posterior indicating probability of pass (grayscale), the GP-derived 50% performance isocontour (grayscale) and the CM $\psi_\theta$ at $K$ = 3 (purple dots).

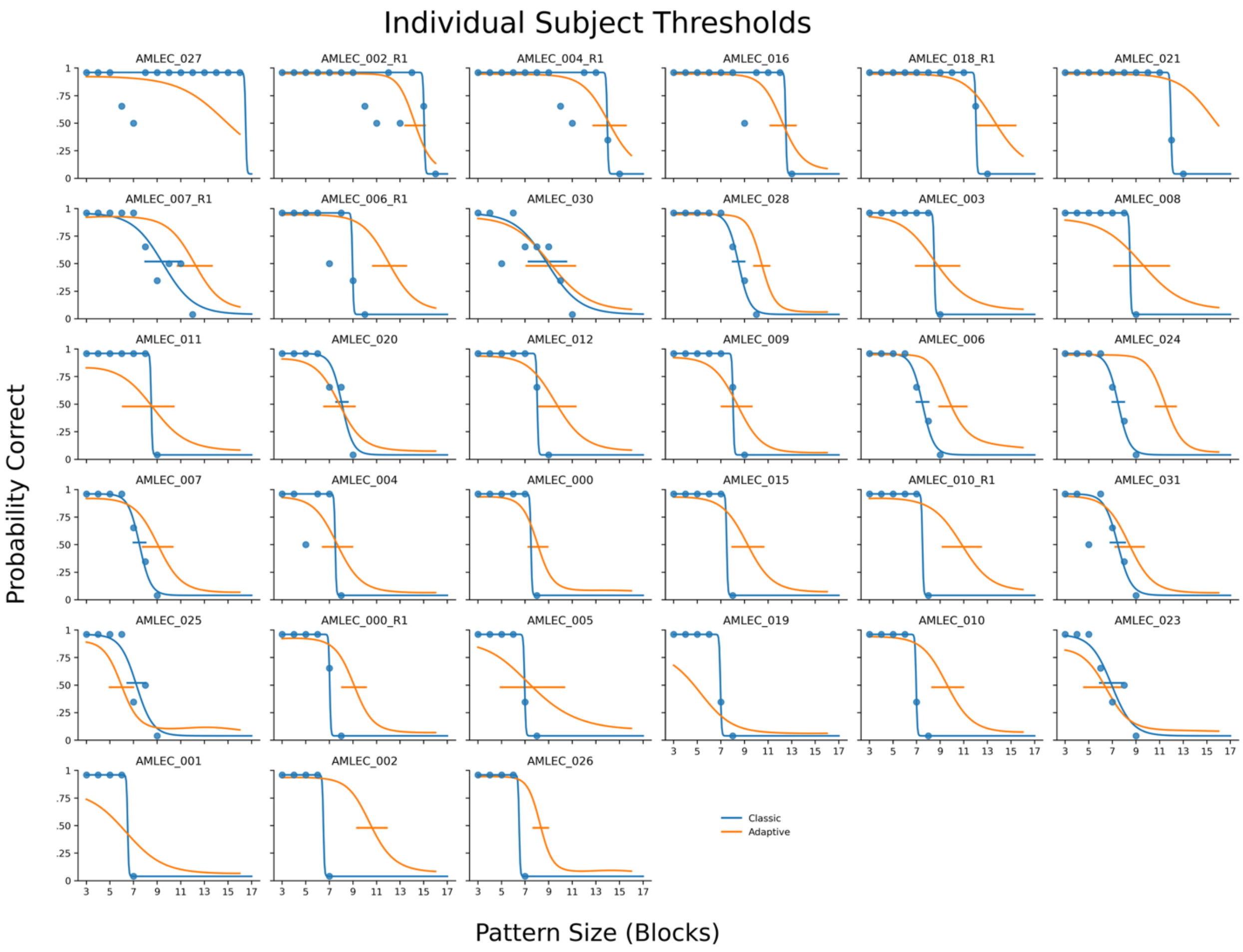

**Figure S2.2**: Estimated psychometric curves at $K$ = 3 with a fixed budget (30 samples) for simulated sessions in the same sort order as Figure S1.

# 3 Reanalysis with Outliers Included

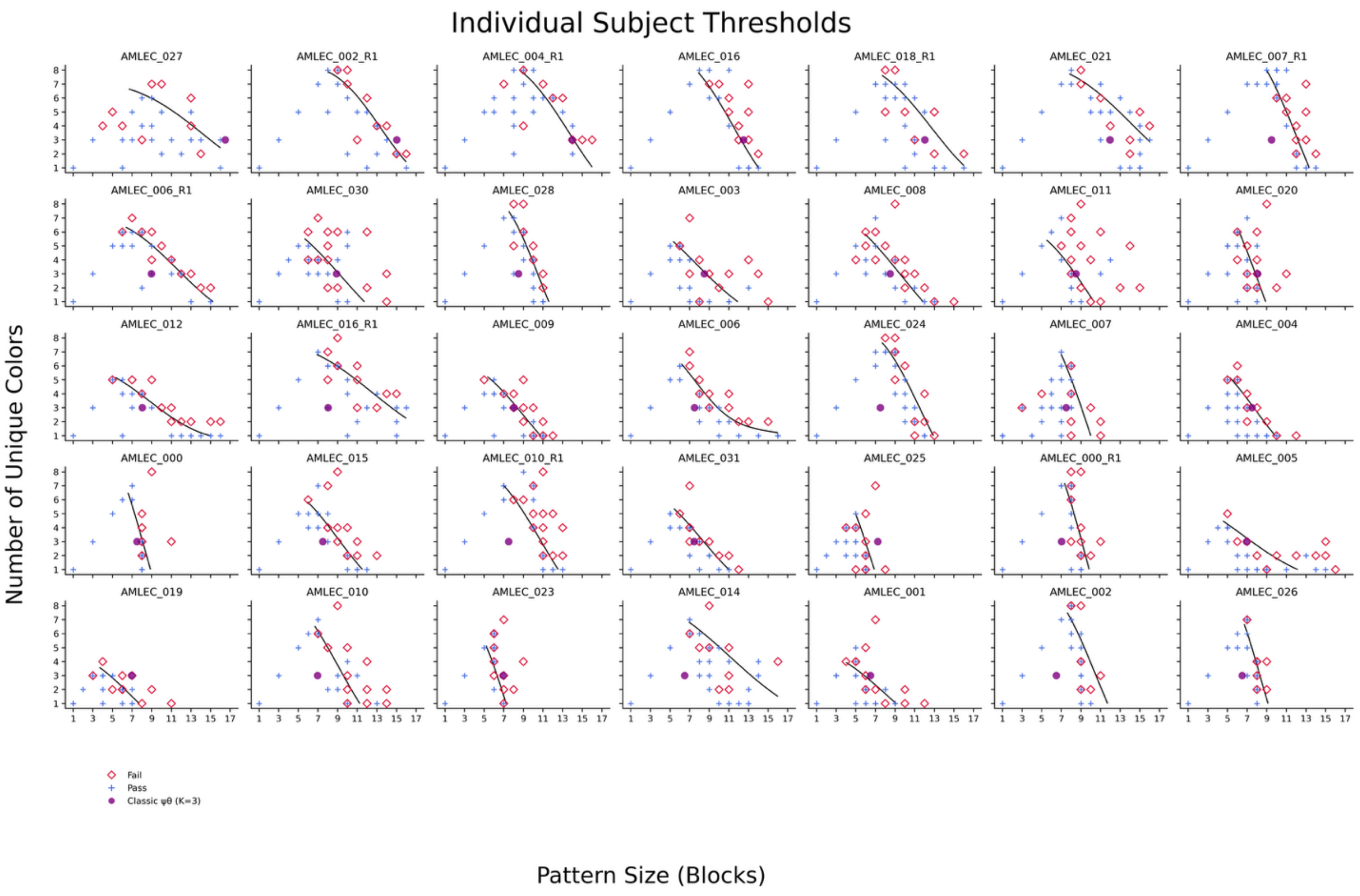

**Figure S3.3.** Two-dimensional performance domains for individual sessions sorted by descending CM $\psi_\theta$. Plotted are AM samples (pass = blue crosses, fail = red diamonds), the GP-derived 50% performance isocontour (black lines) and the CM $\psi_\theta$ at $K$ = 3 (purple dots).

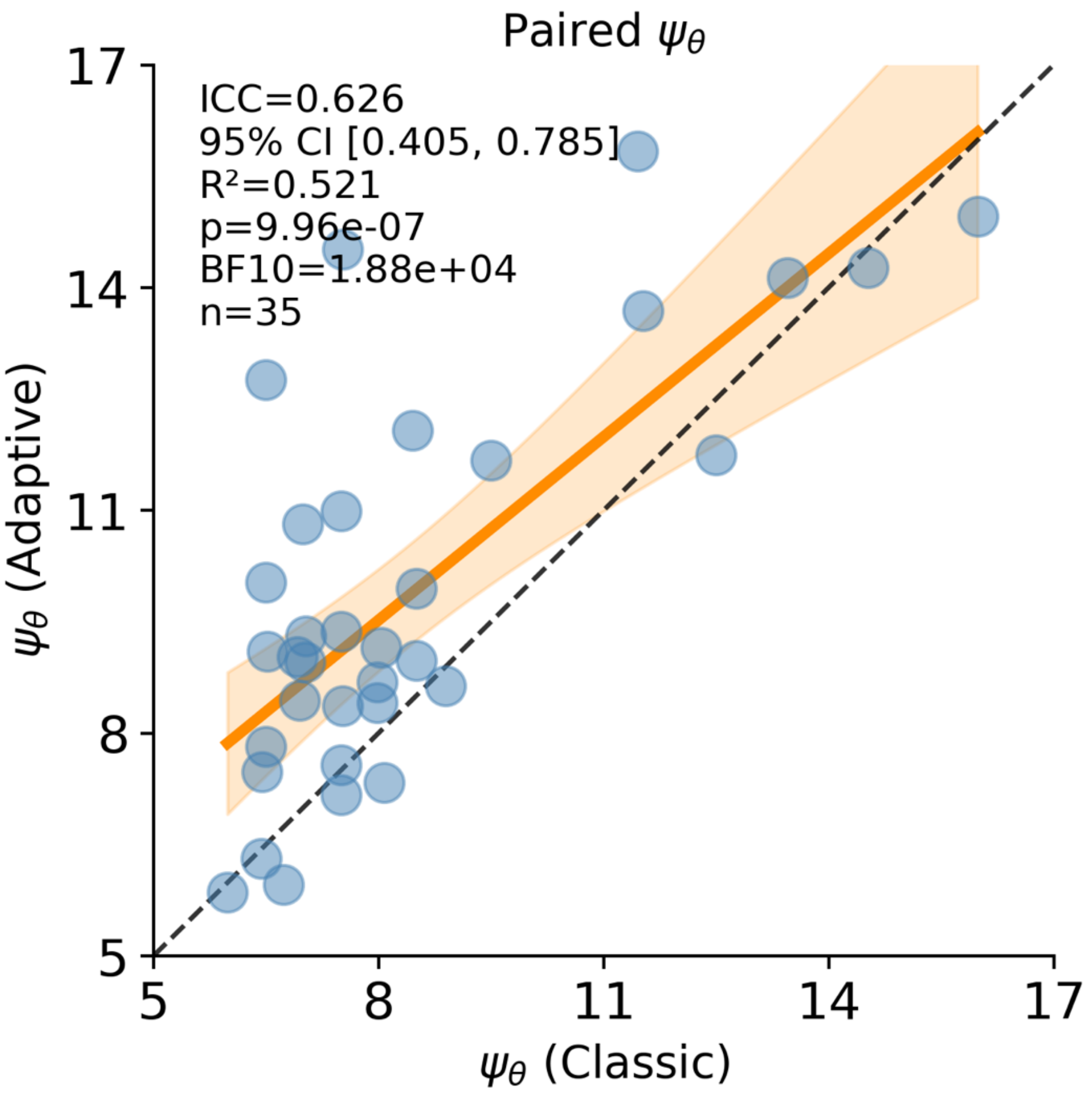

**Figure S3.4.** Agreement between Adaptive and Classic $\psi_\theta$ at $K$ = 3. Each point is one session. The dashed line marks identity; the solid orange line is the linear fit with 95% CI. Values printed on the panel give agreement statistics (ICC(2,1), $p$ value, $BF_{10}$), coefficient of determination $R^2$, and sample size. $\psi_\theta$ is the 50% point of a fitted sigmoid for CM and the 50% point of the GP classifier's predictive posterior for AM.

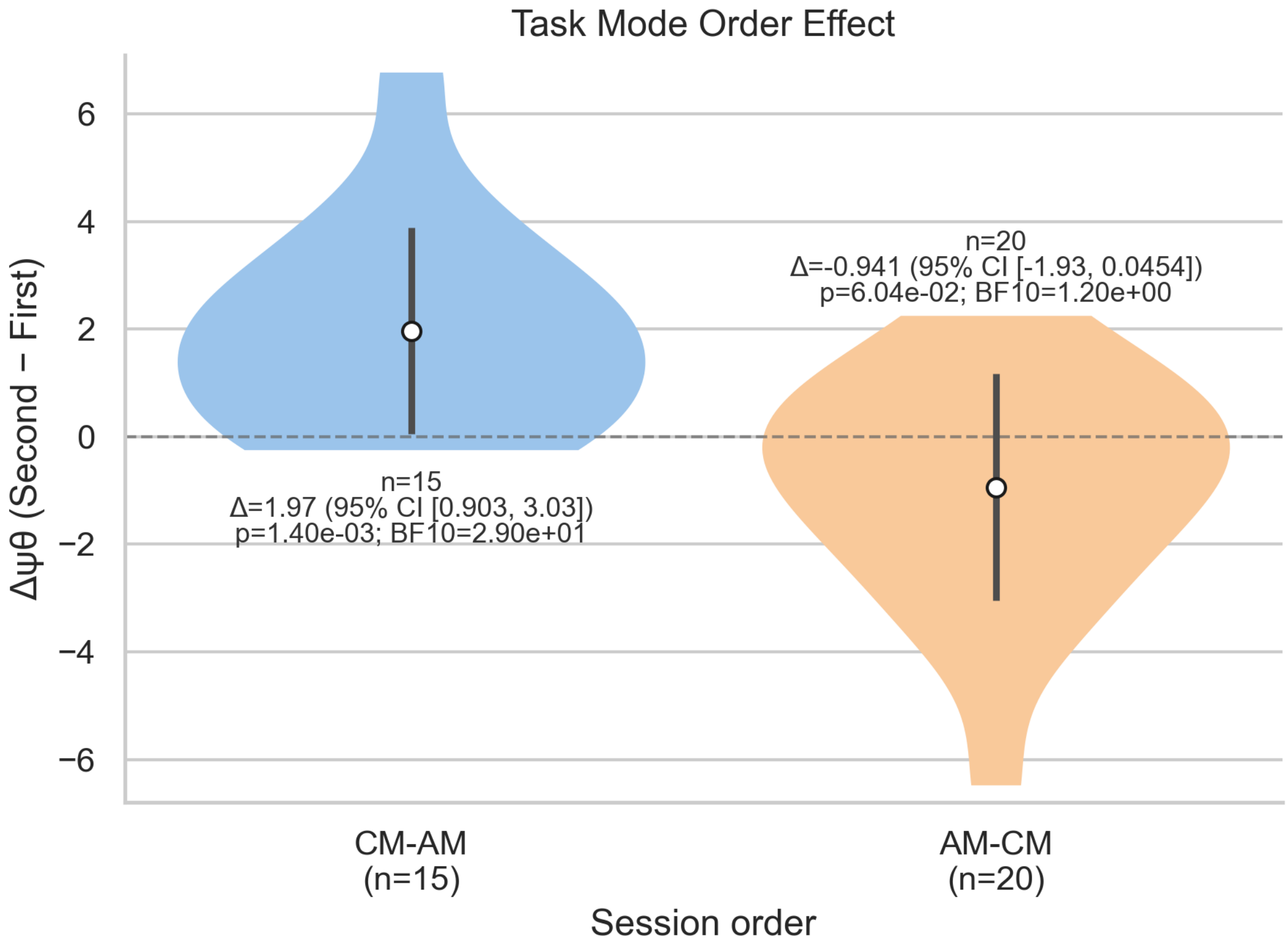

**Figure S3.5.** Session-order effect on $\psi_\theta$. The y axis shows (Session 2 $\psi_\theta$ − Session 1 $\psi_\theta$), split by mode order (CM followed by AM, AM followed by CM). Medians and *p*-values are given beneath each distribution. Sessions that began with CM showed significant improvement on their second mode (AM) (median +1.73), whereas the opposite order revealed no significant change.

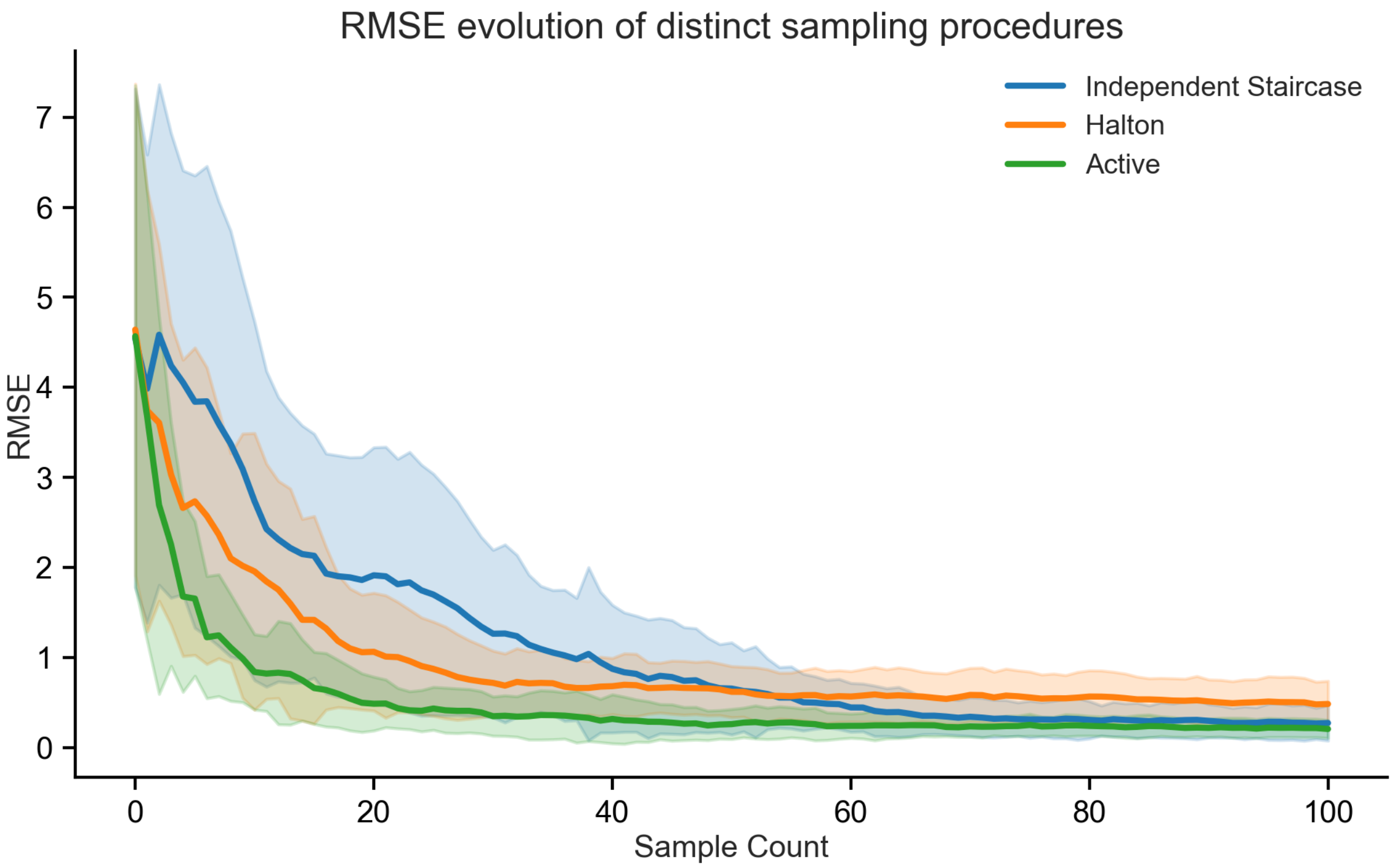

**Figure S3.6**. Root Mean Square Error (RMSE) versus sample count for simulated data. Using 35 session-specific generators, we simulated three sampling policies after a two-trial primer: Independent Staircase (blue), Halton (orange) and Active entropy-maximization (green). Curves show mean RMSE ±1 SD across 33 sessions as the GP is re-fit after each additional observation (up to 100).

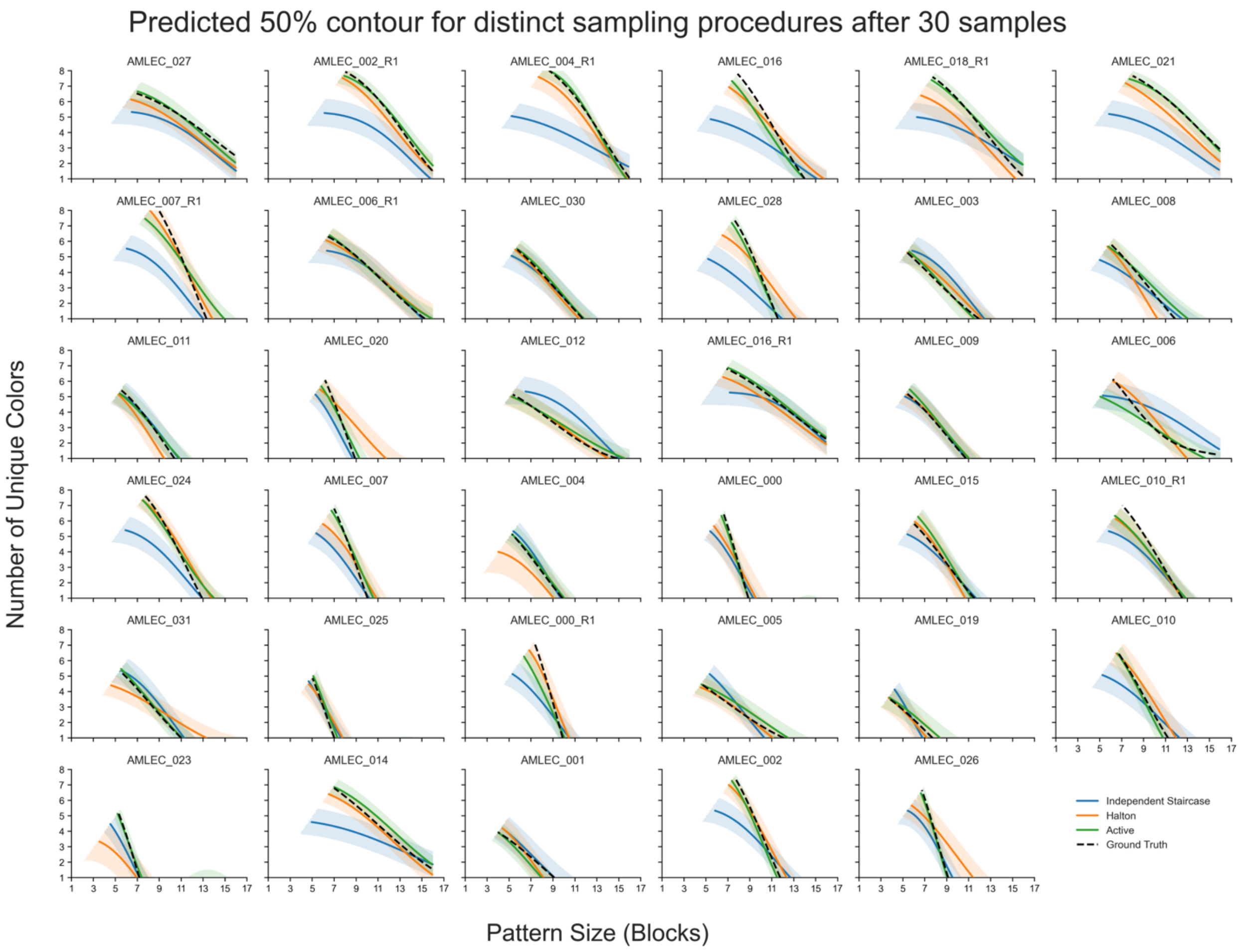

**Figure S3.7**. Contour quality at a fixed budget (30 samples) for simulated sessions in the same sort order as Figure 3. For each session, the dashed curve is the ground-truth 50% isocontour; solid curves are the estimated 50% isocontours from Independent Staircase (blue), Halton (orange) and Active (green). Shaded regions depict each method's ±20% posterior band (0.3–0.7).

# 4 Reanalysis with Repeats Excluded

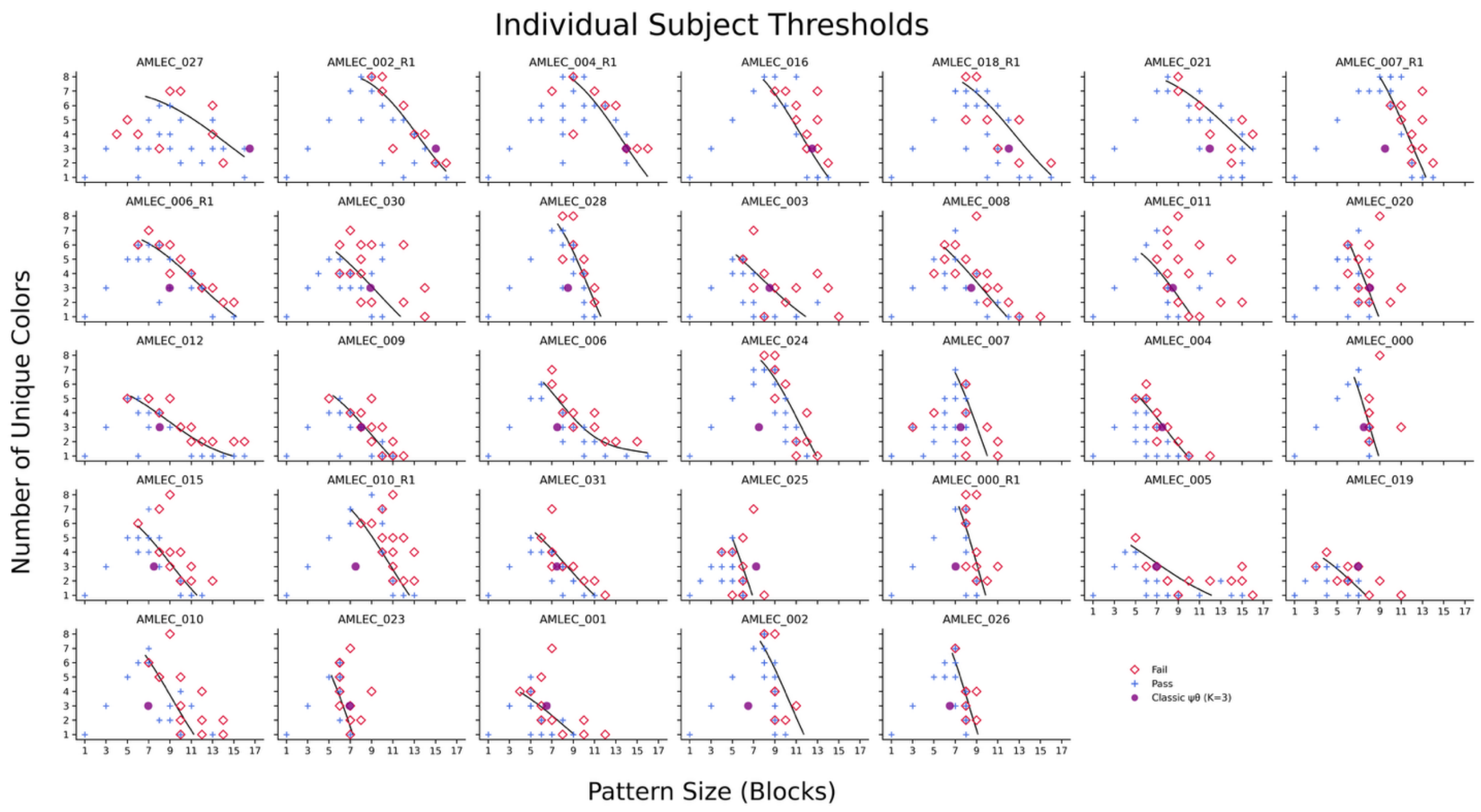

**Figure S4.3.** Two-dimensional performance domains for individual sessions sorted by descending CM $\psi_\theta$. Plotted are AM samples (pass = blue crosses, fail = red diamonds), the GP-derived 50% performance isocontour (black lines) and the CM $\psi_\theta$ at $K$ = 3 (purple dots).

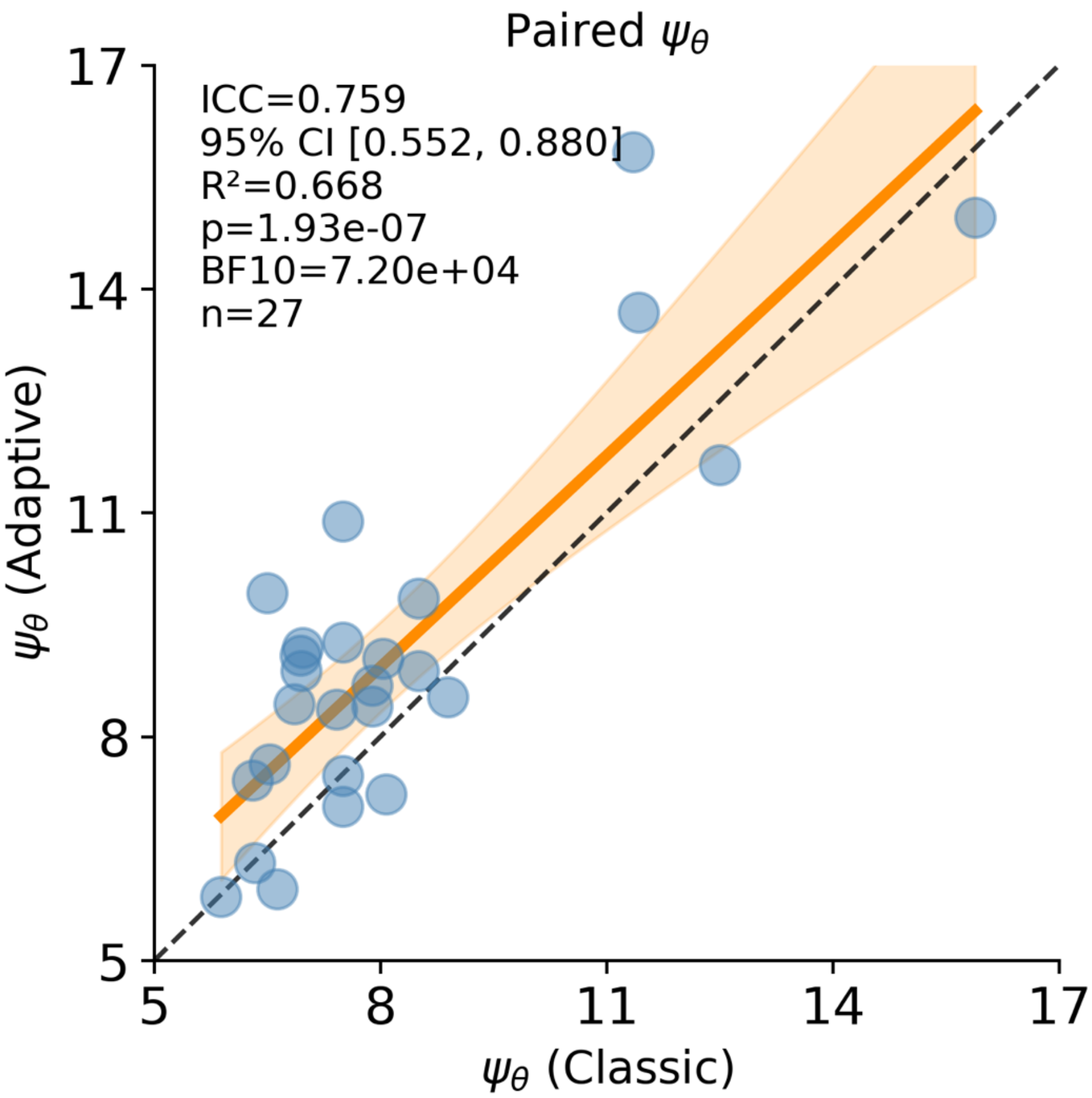

**Figure S4.4.** Agreement between Adaptive and Classic $\psi_\theta$ at $K$ = 3. Each point is one session. The dashed line marks identity; the solid orange line is the linear fit with 95% CI. Values printed on the panel give agreement statistics (ICC(2,1), $p$ value, $BF_{10}$), coefficient of determination $R^2$, and sample size. $\psi_\theta$ is the 50% point of a fitted sigmoid for CM and the 50% point of the GP classifier's predictive posterior for AM.

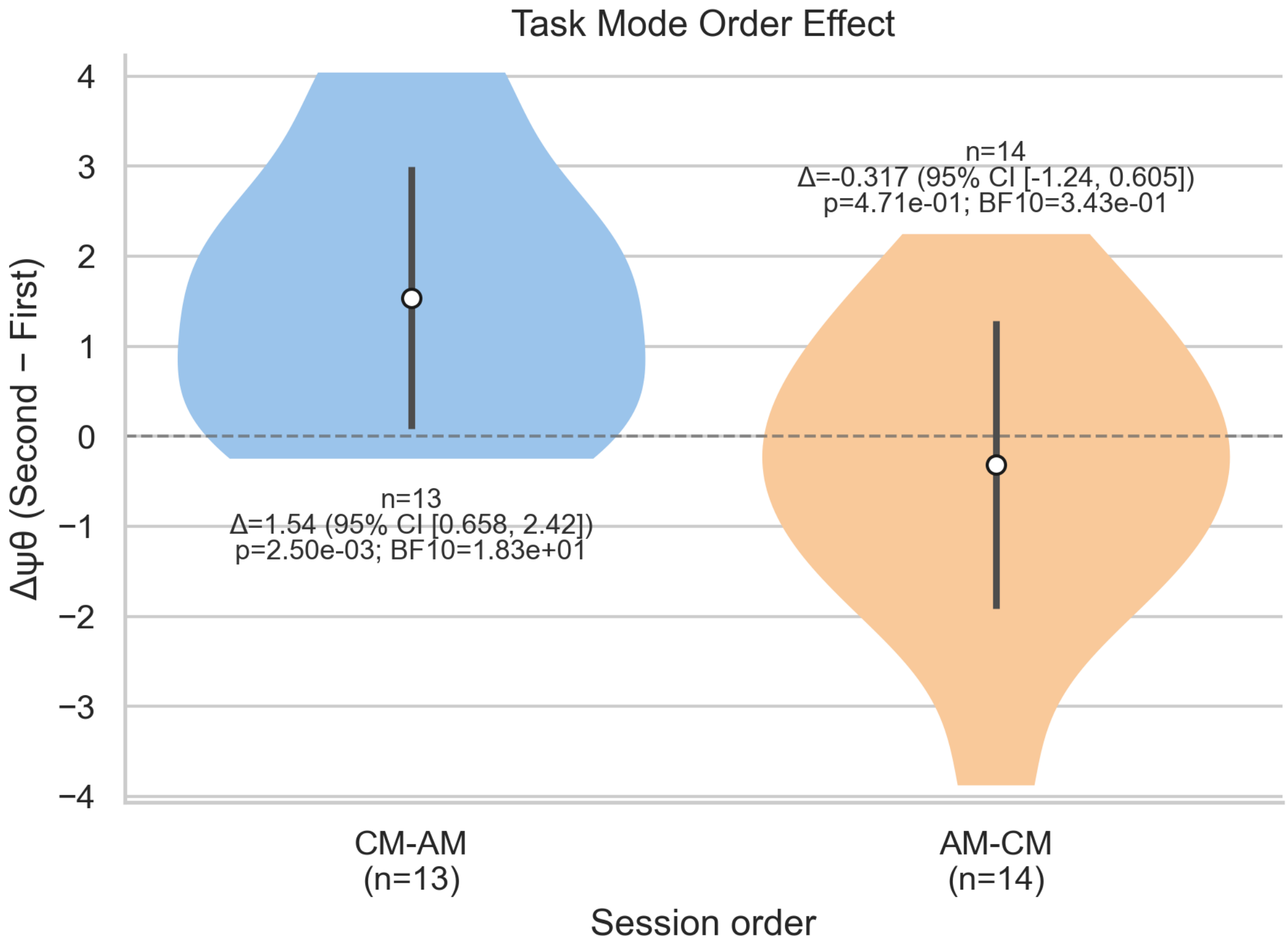

**Figure S4.5.** Session-order effect on $\psi_\theta$. The y axis shows (Session 2 $\psi_\theta$ − Session 1 $\psi_\theta$), split by mode order (CM followed by AM, AM followed by CM). Medians and *p*-values are given beneath each distribution. Sessions that began with CM showed significant improvement on their second mode (AM) (median +1.73), whereas the opposite order revealed no significant change.

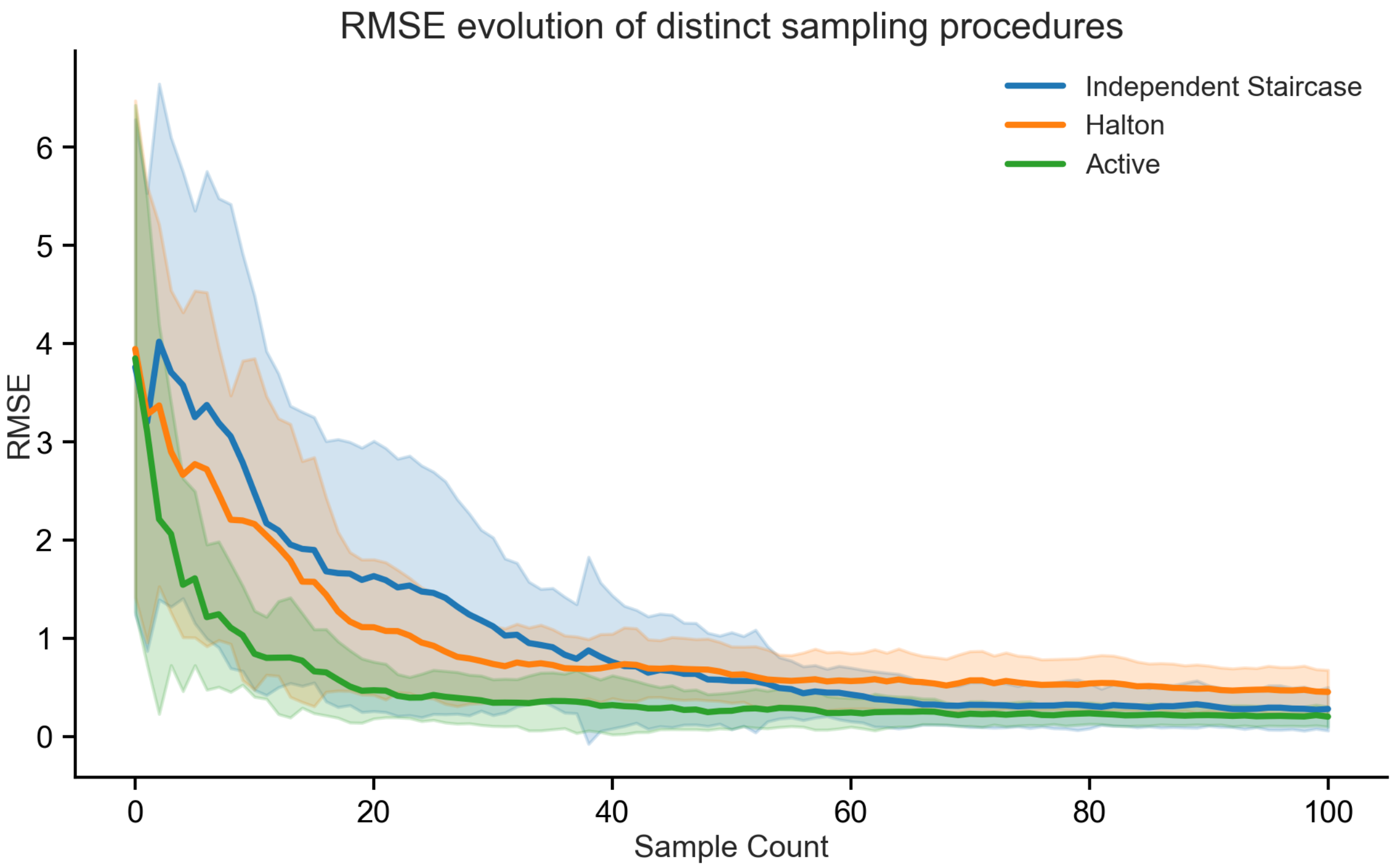

**Figure S4.6**. Root Mean Square Error (RMSE) versus sample count for simulated data. Using 27 session-specific generators, we simulated three sampling policies after a two-trial primer: Independent Staircase (blue), Halton (orange) and Active entropy-maximization (green). Curves show mean RMSE ±1 SD across 33 sessions as the GP is re-fit after each additional observation (up to 100).

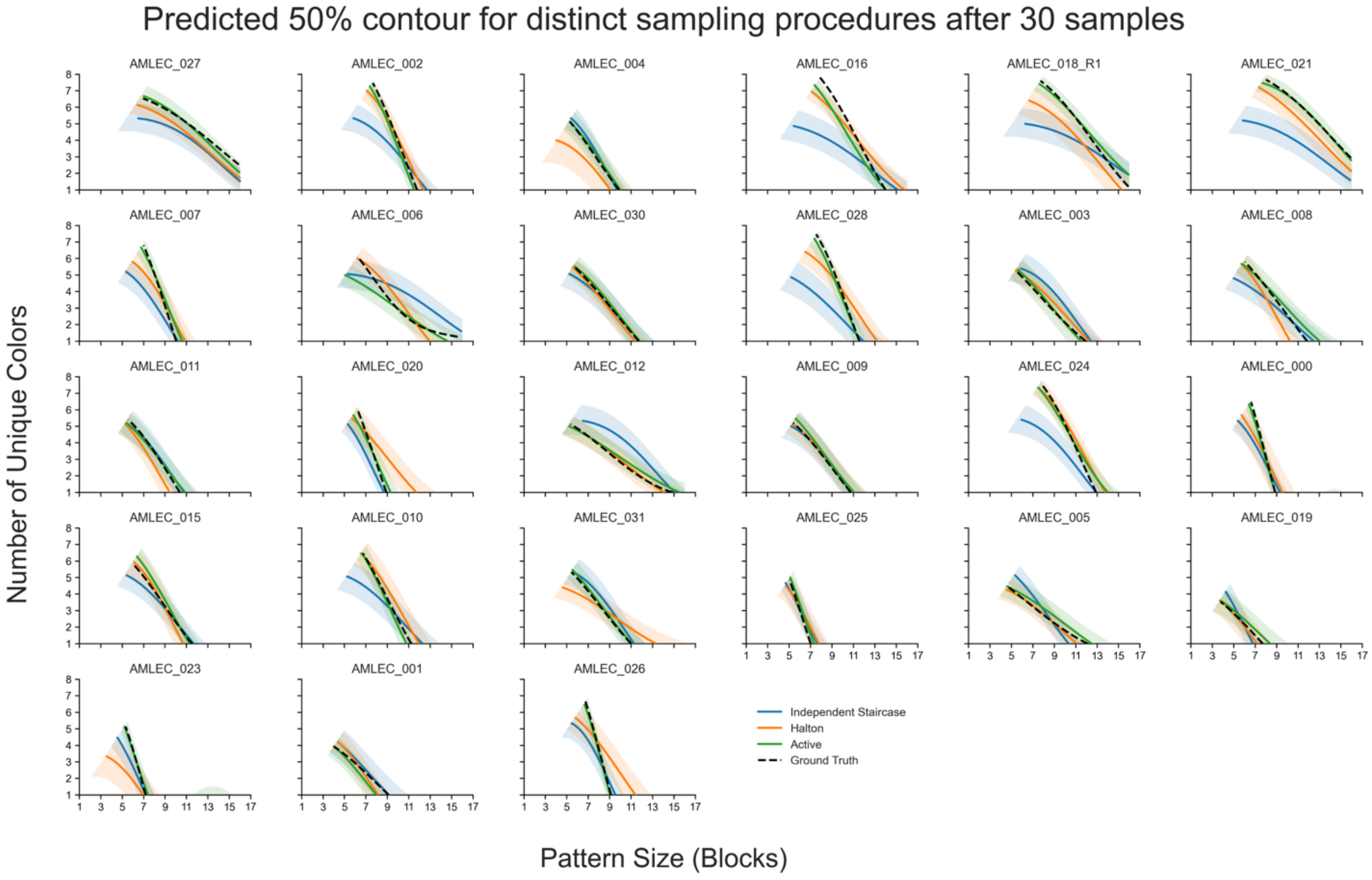

**Figure S4.7**. Contour quality at a fixed budget (30 samples) for simulated sessions in the same sort order as Figure 3. For each session, the dashed curve is the ground-truth 50% isocontour; solid curves are the estimated 50% isocontours from Independent Staircase (blue), Halton (orange) and Active (green). Shaded regions depict each method's ±20% posterior band (0.3–0.7).

# 5 Ghost Point Removal Sensitivity Analysis

The active-learning GP pipeline uses "ghost" (phantom) observations placed at boundary regions of the stimulus domain to regularize the GP posterior. Specifically, ghost points consist of: (1) Template ghosts (failures): points along y = 9 (beyond max colors) and x = 18 (beyond max block size), encoding the prior that very high pattern demands yield failure. (2) Dynamic ghosts (successes): points in the L < K region where the participant is expected to succeed trivially. These ghost points are excluded from candidate selection (the acquisition function never proposes sampling at invalid locations). Their role is solely to regularize the GP posterior at boundaries, preventing overconfidence in unsampled regions.

## Method

To quantify the impact of ghost points on threshold estimation, we refit the GP for each of the 37 sessions in the dataset using only real experimental observations (ghost points removed). All other hyperparameters were held constant (lr = 0.125, β = 1, min lengthscale = 0.15, training iterations = 500, likelihood g = l = 0.05, ARD RBF + linear kernel). We then extracted $\psi_\theta$ (the threshold at K = 3 colors) from both the original (with-ghosts) and refitted (no-ghosts) GP posteriors using the same logistic-fit procedure applied throughout the manuscript.

## Results

Agreement between with-ghost and no-ghost estimates (N = 34 sessions; 3 excluded where original posterior failed to converge): ICC(2,1) = 0.265, Pearson r = 0.572, mean difference (with − without) = −2.41 blocks (SD = 2.80), paired $t(33) = -5.03$, $p = 1.7 \times 10^{-5}$. Sessions hitting parameter cap (17.0) without ghosts: 4/34 (11.8%); with ghosts: 0/34 (0%).

Removing ghost points systematically inflates $\psi_\theta$ estimates by approximately 2.4 blocks on average. Without boundary regularization, the GP posterior in the high-difficulty region becomes unconstrained, causing the logistic fit to saturate at or near the parameter cap.

Convergent validity with the classic method (CM): With ghosts, ICC(2,1) vs CM = 0.526, Pearson r = 0.680. Without ghosts, ICC(2,1) vs CM = −0.055, Pearson r = 0.305. The with-ghost GP estimates show moderate-to-good convergent validity with the independent CM threshold. Removing ghost points worsens convergent validity, confirming that the boundary regularization is essential for producing valid estimates.

## Interpretation

Ghost points serve as an informative prior over boundary behavior — they encode the uncontroversial assumption that performance degrades at extreme difficulty levels and remains high at trivially easy levels. This analysis demonstrates that: (1) Ghost points do not distort threshold estimates — they improve their validity. (2) Without boundary regularization, GP estimates are systematically inflated at the data counts tested and lose correspondence with an independent psychophysical measure. (3) The ghost point strategy is analogous to placing a weakly informative prior on the tails of the psychometric function, a standard practice in Bayesian psychophysics.

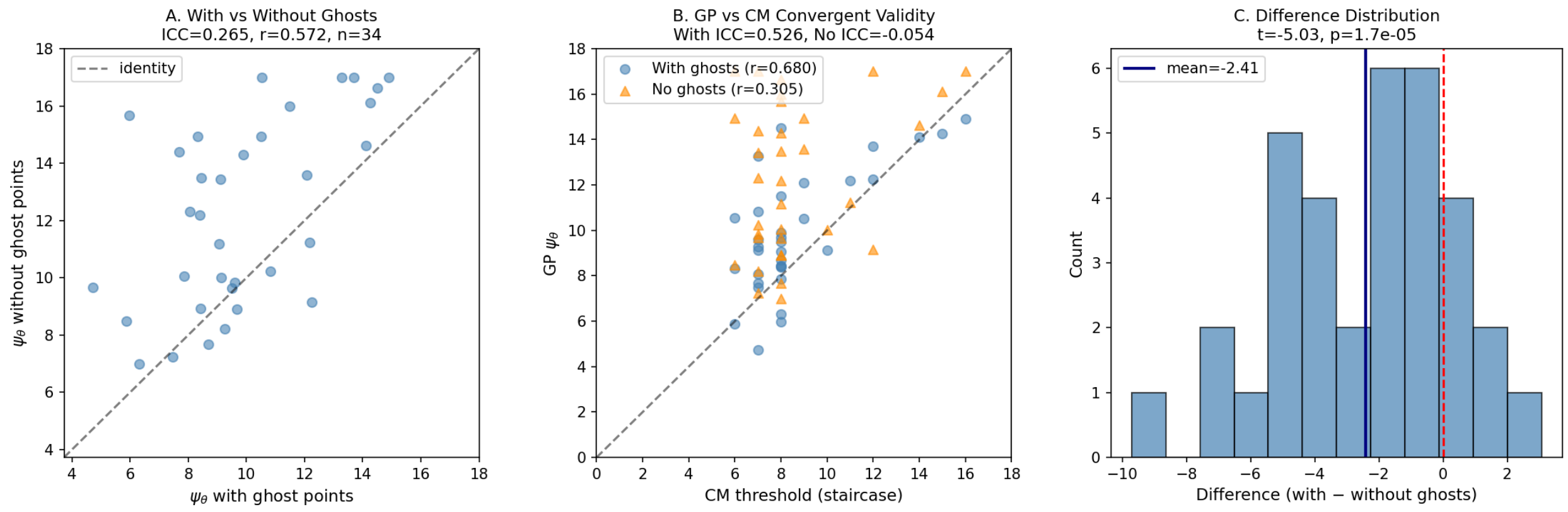

**Figure S5.1.** Sensitivity of GP threshold estimates to ghost point removal. (A) Scatter plot of ψ_θ estimates with versus without ghost points (identity line dashed). (B) Convergent validity: GP estimates against the independent classic-method threshold, separately for with-ghost (circles) and without-ghost (triangles) conditions. (C) Distribution of threshold differences (with − without), showing systematic negative shift when ghosts are removed (mean = −2.41 blocks, $p < 0.001$).

# 6 Supplementary Analyses Addressing Figure 4 Offset

## Trial-Budget Control

To test whether the Figure 4 offset between AM and CM could be explained by unequal total trial counts, we repeated the AM analysis after matching each session to its observed CM trial count. For each of the 33 sessions retained in the Figure 4 cohort, we took the observed CM trial count (n_CM), truncated the AM sequence to its first n_CM trials in the original acquisition order, refit the GP using the same standardized procedure, and re-extracted ψ_θ at K = 3.

The full AM estimate remained positively shifted relative to CM, with a mean signed difference of +1.06 blocks (95% CI [0.48, 1.65]), $t(32) = 3.70$, $p = 8.01 \times 10^{-4}$. After trial matching, the offset was smaller but still positive: the truncated AM estimate exceeded CM by +0.86 blocks on average (95% CI [0.11, 1.62]), $t(32) = 2.32$, $p = .027$. Agreement did not improve under truncation: ICC(2,1) decreased from 0.747 (95% CI [0.561, 0.864]) for full AM vs. CM to 0.653 (95% CI [0.414, 0.810]) for truncated AM vs. CM.

Critically, the reduction in bias from full to truncated AM was modest and not itself reliable. The mean change in signed difference was −0.20 blocks (95% CI [−0.75, 0.35]), $t(32) = -0.73$, $p = .469$. Thus, matching total trial counts attenuated the AM–CM offset slightly, but did not eliminate it, arguing against a simple "AM had more total trials" explanation.

## Locality-Matched GP Reanalysis Around K = 3

Because CM samples exclusively at K = 3, total trial count is only a partial control. A more direct test is whether the AM estimate changes when the GP is forced to rely primarily on observations at or near K = 3. We selected, for each session, the n_CM AM observations nearest to K = 3, ranking trials by proximity (exact K = 3 first, then K = 2/4, then K = 1/5, etc.) while preserving original acquisition order within each distance shell.

This locality-matched control yielded finite thresholds for 30 of the 33 Figure 4 sessions. The mean signed difference from CM fell from +0.97 blocks for full AM (95% CI [0.35, 1.59]; $t(29) = 3.19$, $p = .003$) to +0.38 blocks for the locality-matched GP estimate (95%

CI [−0.18, 0.94]; $t(29) = 1.37$, $p = .180$). Agreement improved from ICC(2,1) = 0.770 (95% CI [0.582, 0.882]) to ICC(2,1) = 0.832 (95% CI [0.678, 0.916]).

The shift from full AM to the locality-matched GP estimate was itself reliable: mean change −0.59 blocks (95% CI [−0.90, −0.29]), $t(29) = -4.00$, $p = 4.01 \times 10^{-4}$. This is the most direct answer to the smoothing question: when the AM GP is restricted to local information around $K = 3$, the Figure 4 offset is substantially reduced.

## Crossed Model and Sampling Reanalysis

We separated model choice from sampling strategy by estimating the two missing cells of the 2 × 2 design: GP fit to CM staircase data (GP-on-CM) and logistic fit to the subset of AM observations collected at $K = 3$ (Logistic-on-AM). If GP fitting alone were the main source of the Figure 4 offset, then GP-on-CM should reproduce a similar positive shift relative to the original CM estimate.

It did not. GP-on-CM remained very close to CM-Logistic, with ICC(2,1) = 0.989 (95% CI [0.977, 0.994]) and a mean signed difference of only +0.09 blocks. By contrast, fitting the logistic model only to AM trials collected at $K = 3$ showed a mean signed difference of only +0.10 blocks relative to CM-Logistic, but agreement was poor (ICC(2,1) = 0.462, 95% CI [0.141, 0.694]) due to the sparse sampling at exact $K = 3$ (mean 5.48 trials, range 2–10).

Taken together, these analyses suggest that the Figure 4 discrepancy is not solely explained by the GP model in isolation, and is not solely explained by AM's larger total trial count. The locality-matched GP reanalysis identifies cross-K information borrowing as the primary contributor to the offset.